\documentclass[sn-mathphys,Numbered]{sn-jnl}

\usepackage{graphicx}%
\usepackage{multirow}%
\usepackage{amsmath,amssymb,amsfonts}%
\usepackage{amsthm}%
\usepackage{mathrsfs}%
\usepackage[title]{appendix}%
\usepackage{xcolor}%
\usepackage{textcomp}%
\usepackage{manyfoot}%
\usepackage{booktabs}%
\usepackage{algorithm}%
\usepackage{algorithmicx}%
\usepackage{algpseudocode}%
\usepackage{listings}%

\usepackage{geometry} 
\usepackage{caption}
\usepackage{subcaption}
\usepackage{graphicx}

\usepackage{pdflscape}

\usepackage{longtable}

\usepackage{booktabs} 
\usepackage{array} 
\usepackage{url} 

\usepackage{float}

\usepackage[firstpage]{draftwatermark}
\SetWatermarkText{DRAFT \\ 2025-02-18}
\SetWatermarkScale{5}
\SetWatermarkLightness{0.8}

\usepackage{flafter}

\usepackage{wasysym}

\usepackage{fancyhdr} 

\hypersetup{
    colorlinks=true,
    linkcolor=blue,
    citecolor=blue,
    urlcolor=blue,
    bookmarks=true,
    bookmarksnumbered=true,
    pdfstartview=FitH
}

\theoremstyle{thmstyleone}%
\theoremstyle{thmstyletwo}%

\theoremstyle{thmstylethree}%

\begin{document}

\title[Multimodal feature extraction for reproducible science]{Nyxus: A Next Generation Image Feature Extraction Library for the Big Data and AI Era}


\author[1,2,3]{\fnm{Nicholas} \sur{Schaub}$^{\dagger}$}

\author[1]{\fnm{Andriy} \sur{Kharchenko}$^{\dagger}$}

\author[1,3]{\fnm{Hamdah} \sur{Abbasi}}

\author[1,3]{\fnm{Sameeul} \sur{Samee}}

\author[1,3]{\fnm{Hythem} \sur{Sidky}}

\author*[1,2,3]{\fnm{Nathan} \sur{Hotaling}}\email{Nathan.Hotaling@axleinfo.com}

\affil*[1]{\orgdiv{DSBU}, \orgname{Axle Research}, \orgaddress{\street{6116 Executive Blvd Suite 400}, \city{Rockville}, \postcode{20852}, \state{MD}, \country{USA}}}

\affil[2]{\orgdiv{NGRF}, \orgname{NovaGen Research Fund}, \orgaddress{\street{6116 Executive Blvd Suite 400}, \city{Rockville}, \postcode{20852}, \state{MD}, \country{USA}}}

\affil[3]{\orgname{NCATS}, \orgaddress{\street{9800 Medical Center Dr}, \city{Rockville}, \postcode{20850}, \state{MD}, \country{USA}}}


\abstract{
Modern imaging instruments can produce terabytes to petabytes of data for a single experiment. The biggest barrier to processing big image datasets has been computational, where image analysis algorithms often lack the efficiency needed to process such large datasets or make tradeoffs in robustness and accuracy. Deep learning algorithms have vastly improved the accuracy of the first step in an analysis workflow (region segmentation), but the expansion of domain specific feature extraction libraries across scientific disciplines has made it difficult to compare the performance and accuracy of extracted features. To address these needs, we developed a novel feature extraction library called Nyxus. Nyxus is designed from the ground up for scalable out-of-core feature extraction for 2D and 3D image data and rigorously tested against established standards. The comprehensive feature set of Nyxus covers multiple biomedical domains including radiomics and cellular analysis, and is designed for computational scalability across CPUs and GPUs. Nyxus has been packaged to be accessible to users of various skill sets and needs: as a Python package for code developers, a command line tool, as a Napari plugin for low to no-code users or users that want to visualize results, and as an Open Container Initiative (OCI) compliant container that can be used in cloud or super-computing workflows aimed at processing large data sets. Further, Nyxus enables a new methodological approach to feature extraction allowing for programmatic tuning of many features sets for optimal computational efficiency or coverage for use in novel machine learning and deep learning applications.
}




\maketitle

\noindent$^{\dagger}$These authors contributed equally to this work.

\section{Main}\label{secMAIN}

The size and complexity of scientific and clinical image data is growing rapidly.\cite{BrayCarpenter2018} Advances in scientific and clinical imaging equipment are producing images with higher spatial resolution,\cite{KashaniCharacterization2023,DamstraVisualizing2022} higher dimensionality (i.e. time and volume),\cite{LiuHarnessing2021} multiplexing and hyperplexing (i.e. more spectral channels),\cite{MaricWhole2021} coupled with greater automation (i.e. more replicates and sample conditions).\cite{DengScalable2023,KunduHigh2022} These improvements in equipment and automation are unlocking significantly larger experimental spaces, leading to massive increases in imaging dataset sizes (petabytes and beyond). Methods designed to extract measurements from images in the  past were designed for significantly smaller scales of data, and as a result often lack one or several of the following capabilities: 
\begin{enumerate}
    \item All measurements can be made on images too large to fit into memory (improving reproducibility of big image analysis by organizations with less compute resources and enabling low resource communities to access state-of-the-art analysis algorithms).
    \item Configurable and tunable feature hyperparameters (enabling different communities to set their profiles, or for novel profiles such as AI/ML use cases to be optimized for).
    \item Highly optimized feature algorithms to dramatically increase measurement speed at scale.
    \item Utilization of hardware acceleration (Graphical Processing Units – GPU).
    \item Modular software architecture to ease new feature addition from the community (developers only need to understand their algorithm, not all of software infrastructure).
    \item Diverse user interface availability making the method highly accessible for many use cases and user skill levels (available in common scripting language package managers (e.g. PyPI, Conda, etc.), Open Container Initiative (OCI) compliant packaging, command line interface (CLI), popular workflow language tools (e.g. Common Workflow Language - CWL - tool), and as a graphical user interface).
\end{enumerate}

Image feature extraction is the process of deriving quantitative measurements from images. Comprehensive feature extraction in the clinical imaging domain is often referred to as radiomic profiling\cite{TomaszewskiBiological2021} and in the cellular microscopy domain as cell profiling.\cite{StirlingCellprofiler2021} These communities have developed extensive feature sets in parallel, and they can be broadly organized into five groups: morphological/shape, intensity, texture, volumetric, and miscellaneous. Across the radiomic and cell profiling domains there are generally two broad feature extraction use cases: (1) image features, which are features extracted from entire images (stitched together or individual acquisitions) and (2) regional features, or features extracted from regions of interest (ROIs) or masks within images. Image features are often used for assessments of image quality (e.g. blurriness, motion, contrast, etc.) but have also been shown to be useful in classification and predictive tasks.\cite{uhlmann_cp-charm:_2016} The second use case is used for a wide variety of applications including biomarker identification, classification, and regression tasks. Despite the prevalence of these two use cases, current feature extraction methods are not designed to support extracting all supported features for both use cases. Most software libraries are optimized for one use case or another, or have optimized execution of a subset of features but not others. Further,  feature extraction libraries often have custom implementations of identical features (e.g. perimeter) which leads to disparity in measurements across libraries. These disparities are especially apparent between the radiomic and cell profiling fields, where the two communities have not coordinated their implementations or standards and thus often produce different measures for the same feature.\cite{bajcsy_etal_2021, bettinelli_novel_2022} Given the above, a next generation feature‑extraction library is needed that supports both use cases, regardless of image dimensions, pixel‑intensity distributions or feature groups. It should incorporate common standards and methods from both the radiomic and cell‑profiling communities to bridge their divide, reduce parallel development, and increase the replicability and reproducibility of research.

Here we present Nyxus, a feature extraction library focused on robust, scalable image feature extraction algorithms to support the analysis of big image data. Nyxus was designed from the ground up to address the issues above as well as many other challenges encountered with current feature extraction methods: execution parallelism,  memory safety, configurable hardware resource utilization, logging, robust error handling, and/or extensive documentation of measurements and their implementation. Nyxus was designed to support current and future big data and AI needs but does not sacrifice on the strong measurement foundation of the past.  The feature metrics used in Nyxus are rigorously tested for accuracy and precision against standardized datasets.\cite{zwanenburg_image_2020,bettinelli_novel_2022} We also compare Nyxus to other tools used for clinical and scientific image feature extraction including PyRadiomics, MITK,\cite{Goch_Metzger_Nolden_MITK_2017,Gotz_etal_mitk_phenotyping_2019} and RadiomicsJ\cite{kobayashi_2022} for clinical imaging, and CellProfiler,\cite{McQuin_Goodman_Chernyshev_etal_Cellprofiler_2018} Imea,\cite{Kroell_Imea_2021} NIST WIPP plugin,\cite{nist_wipp_plugin} and WND-CHARM\cite{Orlov_etal_WNDCHARM_2008} for cell profiling. We also compare to MatLab's default image feature extraction library to compare to a domain agnostic library with a known focus on numerical accuracy of its measurements. A broad comparison of feature offerings of Nyxus as compared to other open libraries is provided in Figure \ref{fig:FUNCTIONALITY_HEATMAP} and in Supplementary Tables 1-5. Nyxus's source code is publicly available and licensed under the highly permissive MIT license for use in public and private applications.\cite{polus_nyxus_opensource}

%
%

\begin{figure}[h!]
    \centering
    \includegraphics[trim=1cm 1cm 1cm 1cm, clip, width=\textwidth]{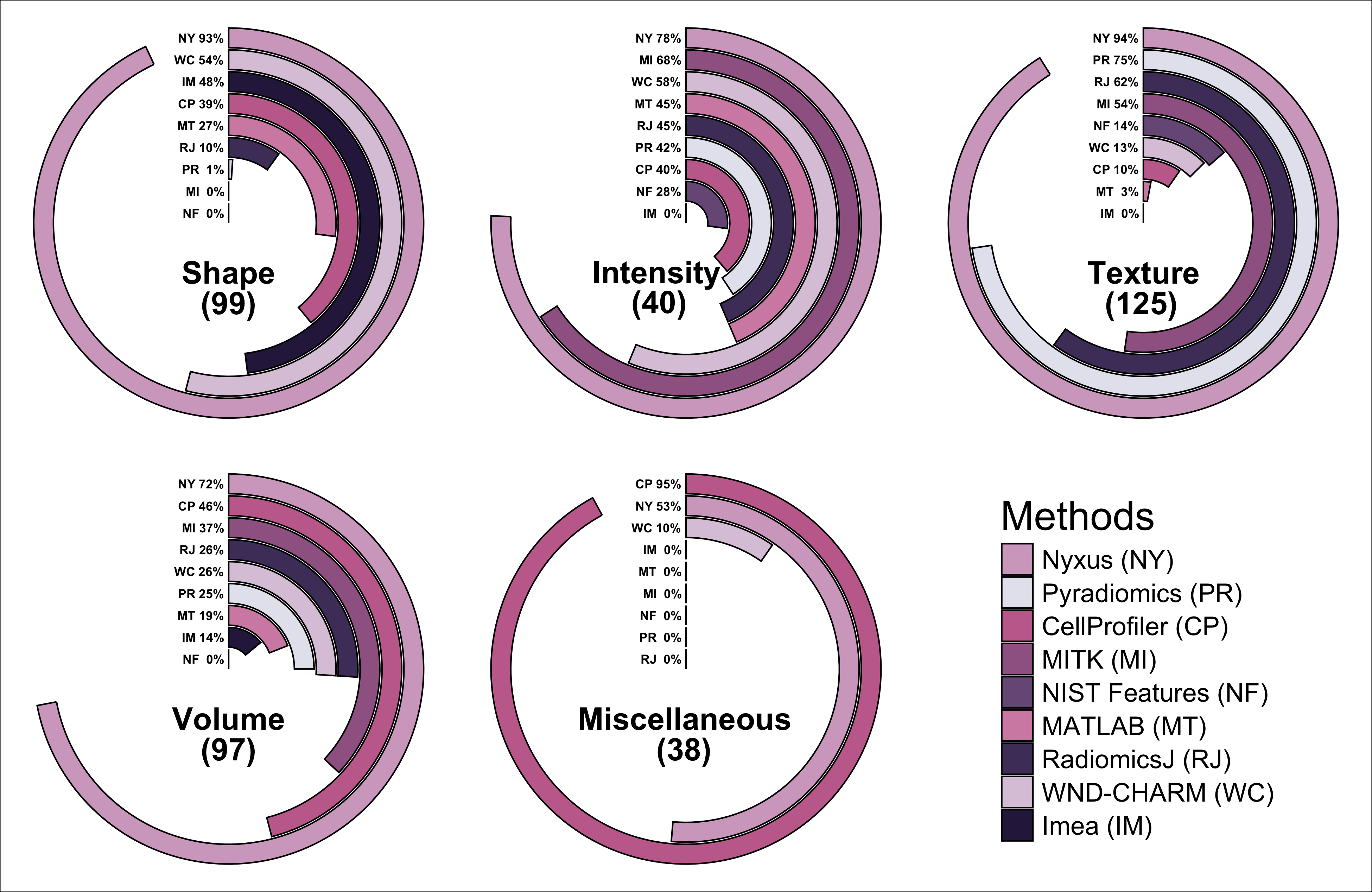}
    
    \caption{
Nyxus feature set compared to other common image feature extraction Methods. Each racetrack plot compares features from 9 different feature extraction libraries; NY - Nyxus, WC – WND-CHRM, RJ- RadiomicsJ, PR- PyRadiomics, NF- NIST Java Features, MI- MITK, MT- MatLab, IM- Imea, CP- CellProfiler.\cite{McQuin_Goodman_Chernyshev_etal_Cellprofiler_2018} Each plot compares features classified as shape, intensity, texture, volumetric, and miscellaneous.  Complete information on individual features can be found in Supplemental Tables \ref{table:INTEN} (Intensity), \ref{table:SHAPE_MOMS} (Shape and Moments), \ref{table:TEXTURE} (Texture), \ref{table:MISC} (Miscellaneous), and \ref{table:VOL} (Volumetric).
	}
    
    \label{fig:feature_coverage}
\end{figure}



\newcommand \wid {.02}
\newcommand \hmapwid {6px}
\newcommand \hsp {-3mm}
\newcommand \vsp {-1.5mm}
\newcommand \fontselector {\tiny}	

\begin{figure}[h!]
 

	\centering
    \includegraphics[width=\textwidth,trim=270 270 270 270,clip]{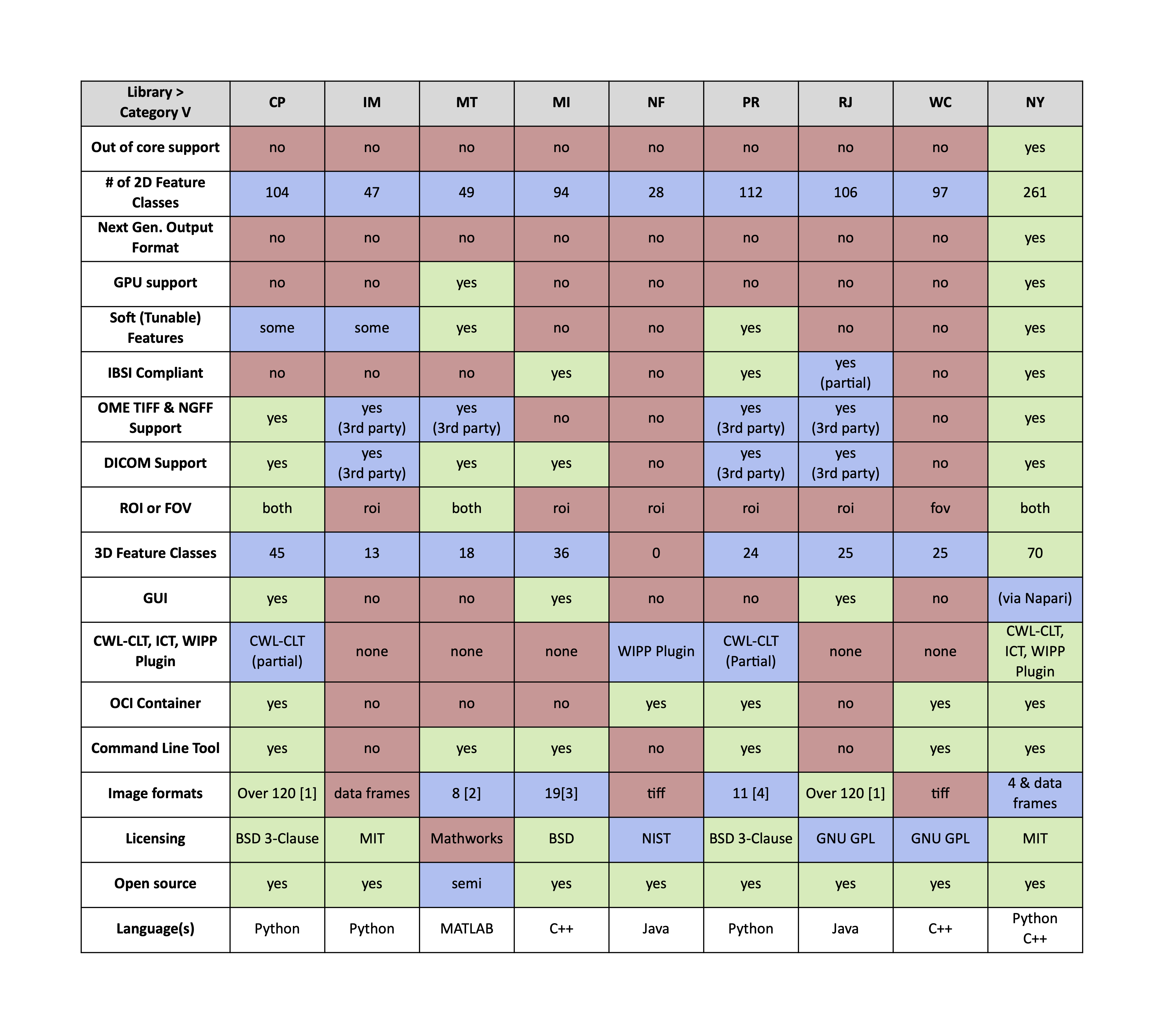}

\caption{Heatmap comparing feature extraction library capabilities across multiple criteria. Color coding is row-specific: green generally indicates full support or optimal capability (e.g., $>90\%$ for quantitative metrics, most permissive for licensing), blue indicates partial support or intermediate capability (e.g., $10\%-90\%$ for quantitative metrics, some restrictions for qualitative metrics), and red indicates minimal/no support or most restrictive (e.g., $<10\%$ for quantitative metrics, proprietary licensing). Supplemental Table \ref{table:FUNC_HEATMAP_CRITERIA} provides detailed definitions for how each row's criteria were classified and what the color coding means in that specific context}

\label{fig:FUNCTIONALITY_HEATMAP} 
\end{figure} 

\section{Results}\label{Results}

\subsection{Scalability}

\begin{figure}[p]
	
    
\centering
\begin{subfigure}[t]{0.49\textwidth}
\centering
\includegraphics[width=6.5cm]{grafix/fgroupwise_timing_tissuenet_UNtargeted_TRIMMED.png}
\caption{Tissuenet 1.0 (Nyxus untargeted)}
\label{figPERF:TISS_UNT}
\end{subfigure}%
~
\begin{subfigure}[t]{0.485\textwidth}
\centering
\includegraphics[width=6.5cm]{grafix/fgroupwise_timing_tissuenet_targeted_TRIMMED.png}
\caption{Tissuenet 1.0 (Nyxus targeted)}
\label{figPERF:TISS_TARG}
\end{subfigure}


\begin{subfigure}[t]{0.485\textwidth}
\centering
\includegraphics[width=6.5cm]{grafix/fgroupwise_timing_decathlon_untargeted_TRIMMED.png}
\caption{Decathlon (Nyxus untargeted)}
\label{figPERF:DECATL_UNT}
\end{subfigure}
~
\begin{subfigure}[t]{0.49\textwidth}
\centering
\includegraphics[width=6.5cm]{grafix/fgroupwise_timing_decathlon_targeted_TRIMMED.png}
\caption{Decathlon (Nyxus targeted)}
\label{figPERF:DECATL_TARG}
\end{subfigure}


\begin{subfigure}[t]{0.48\textwidth}
\centering
\includegraphics[width=6.5cm]{grafix/timing_multicore_TRIMMED.png}
\caption{Scalability on Different Hardware}
\label{figPERF:CPU}
\end{subfigure}
~
\begin{subfigure}[t]{0.48\textwidth}
\centering
\includegraphics[width=6.5cm]{grafix/timing_scalability_TRIMMED.png}
\caption{Scalability as a Function of Region Size}
\label{figPERF:SCALAB}
\end{subfigure}


\begin{subfigure}[t]{0.8\textwidth}
\centering
\includegraphics[trim=0cm 0cm 0cm 0cm, clip, width=\linewidth]{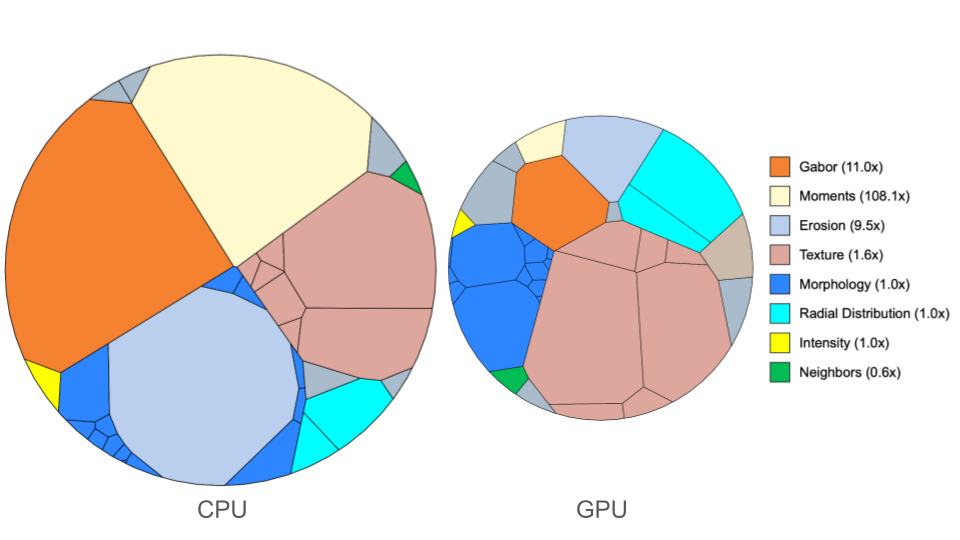}
\caption{CPU (left) versus GPU (right) execution (bigger = slower)}
\label{figPERF:GPUNONGPU}
\end{subfigure}
	
    \caption{Nyxus hardware-specific performance. (a) Tissuenet (Nyxus untargeted), (b) Tissuenet (Nyxus targeted), (c) Decathlon (Nyxus untargeted), (d) Decathlon (Nyxus targeted) - relative Nyxus performance in untargeted and targeted mode versus competition software by feature group on various datasets (Amazon EC2 p5 node). (e) Scalability on different hardware at varying number of CPU threads and various datasets (Notebook hub, EC2, M1). (f) Scalability as a function of region size - CUDA GPU-enabled scalability on synthetic slides having various ROI area and population (Amazon EC2 p5 node using 8 CPU threads). (g) CPU (left) versus GPU (right) execution cost (bigger = slower, sizes are relative to each other) - compute cost structure of featurizing an image of 100x 100K-pixel ROIs on a GPU-disabled and GPU-enabled p5 node. The legend shows color coded broad feature categories, with fold improvement of GPU over CPU in parenthesis.}
    
    \label{fig:PERF}
\end{figure}

To test the capacity of Nyxus to scale to large datasets, feature calculation time was compared across libraries on two standard datasets. The Medical Decathalon \cite{antonelli_medical_2022} dataset, which contains clinically relevant images (fewer but larger ROIs), and the TissueNet \cite{Greenwald_Miller_Moen_etal_2022} dataset, which contains microscopy images of cells and tissues (large numbers of small ROIs). Since Nyxus aims to implement scalable and performant algorithms across all biomedical applications, the performance of Nyxus (and all other libraries) was compared across both of these "typical" use cases to ensure that it outperformed current open-source standards in each domain.  To make as fair of a comparison as possible, calculation times for all features in each library were compared. However, since each tested library contains vastly different numbers and types of features within each Feature Group, and because of the large total number of features to compare (see Supplemental Tables \ref{table:INTEN}-\ref{table:VOL}), the total execution time was measured only for the features that each library supported within a given Feature Group. A key methodological note is that Nyxus has implemented more features than any other library except the miscellaneous category (see Figure \ref{fig:feature_coverage}). This means the timing comparisons shown in Figures \ref{figPERF:TISS_UNT}-\ref{figPERF:DECATL_TARG} are very conservative on the speed increase that Nyxus provides as Nyxus calculates more features in a given Group than its competitors (except miscellaneous). Normalization by the number of features calculated per group was not performed due to the incomparability of feature computational complexity. The raw timing values (in seconds) for all libraries across both datasets are provided in Supplemental Table \ref{table:BENCH}. 
 
Results on TissueNet images (2,498 images total) show Nyxus outperforms tools commonly used in microscopy (Figure \ref{figPERF:TISS_UNT} and Figure \ref{figPERF:TISS_TARG}). Figure \ref{figPERF:TISS_UNT} shows the fold change of feature calculation time (as compared to Nyxus) with default parameter settings in Nyxus, where feature calculations are "untargeted" (feature hyperparameters have not been optimized for speed of calculation). Nyxus is more than $3 \times$ (intensity) to almost $35 \times$ (texture) faster than CellProfiler (CP), a widely used tool for microscopy image analysis. For "targeted" (feature hyperparameters have been optimized for speed of calculation) feature extraction, the performance improvement is $58 \times$ (intensity) to $131 \times$ (texture) faster than CP. Importantly, these fold changes are conservative estimates because each bar in Figure \ref{fig:PERF} has not been normalized by the number of features calculated for a feature group and Nyxus has the largest number of features in each group except miscellaneous (see Figure \ref{fig:feature_coverage}). The results in Figure \ref{fig:PERF}  which show that NIST Feature Extractor (NF) and WND-CHARM (WC) have a faster (less than $1 \times$) execution time than Nyxus, are explained by Nyxus having a dramatically larger number of features (NF = 28 and WC = 122 features, relative to 261 features for Nyxus).

Nyxus also outperformed clinical imaging tools on the TissueNet data. MITK (MI) and RadiomicsJ (RJ) were 100s to 1,000s of times slower than Nyxus on the TissueNet dataset. This is because the structure and format of the data is not what these libraries were designed to handle (i.e. many small ROIs per image). Clinical data commonly contains one ROI per image mask, and processing multiple ROIs for an image requires reading multiple files and processing each ROI separately. Thus, the times shown in Figure \ref{figPERF:TISS_UNT} and Figure \ref{figPERF:TISS_TARG} include converting a multi-ROI image into a single ROI per image mask and loading/extracting features across all these individual images. The time to load and extract features from 100s of individual images with a single ROI per image is considerable and the main reason for this stark performance difference. This was done to get a total time comparison across all Methods of labeled image mask to calculated features. However, the authors acknowledge that this represents a “worst-case” scenario for these libraries and is a use case and data structure that they were not well designed to handle.

Figure \ref{figPERF:DECATL_UNT} and Figure \ref{figPERF:DECATL_TARG} show the same analysis on a clinical dataset (Medical Decathlon) which offers a more fair assessment of the radiometric libraries' performance. As expected, the clinical feature extraction libraries significantly improved their performance on the Medical Decathlon dataset.  However, Nyxus was still more than 5x faster across any feature group than RadiomicsJ with its default settings (Figure \ref{figPERF:DECATL_UNT}) and was almost $2 \times$ faster than MITK on the intensity feature group. While MITK was $1.25 \times$ (0.8) faster than Nyxus on the texture feature group, MITK had just over half as many features as Nyxus: 67 vs 118 ($\sim 56.8\%$) respectively. Similar to Figure \ref{figPERF:TISS_TARG} larger performance gains were seen on the “targeted” or optimized features in Figure \ref{figPERF:DECATL_TARG} between Nyxus and other methods. Leading to Nyxus outperforming all libraries across all feature groups by at least 1.46x (max 357x). Unexpectedly, Pyradiomics (PR) performed far better than its radiometric profiling peers on TissueNet and far worse on the Medical Decathalon dataset, even though it was designed for clinical image feature extraction. The authors hypothesize this is due to how data loading/saving is performed by PR but a root-cause analysis for this difference was outside of the scope of this manuscript. CP was also significantly slower on the Medical Decathalon dataset ($20 \times-198 \times$ slower than Nyxus) than the TissueNet dataset ($3-35 \times$ slower than Nyxus). This was attributed to the fact that it was not designed with this data structure (many “large images” with very few large ROIs) in mind and therefore represents a “worst-case” scenario for this library.

In addition to doing a direct comparison of Nyxus to other feature extraction libraries, we also compared performance across different commonly encountered computational environments. Figure \ref{figPERF:CPU} shows that Nyxus scales well across a variety of computational infrastructures and can be multi-threaded in a variety of node types (linux EC2 instances, Apple Silicon, a Jupyter Notebook Server environment, etc.). On benchmark datasets (such as TissueNet) or synthetic data, Figure \ref{figPERF:CPU} shows that the benefits of threading diminish after 6-10 threads depending on the infrastructure type.  Similar benefits of thread counts are seen no matter the size and count of ROIs tested in synthetic data, but in general larger ROIs with higher counts benefit more from threading than do smaller ROIs with lower counts. The Macbook Pro times also highlight Nyxus’ ability to execute on ARM architectures efficiently, with best in class results shown on TissueNet. Figure \ref{figPERF:CPU} also highlights a unique aspect of image feature analysis in that high performance compute infrastructure (very high thread counts and memory) is not advantageous (after $\sim 10$ threads) to increase performance, and instead image feature extraction benefits more from “embarrassingly parallel” architectures in which individual images/volumes analyses are spun onto processes with relatively few threads and memory in parallel.

Figure \ref{figPERF:SCALAB} plots the effect of total feature extraction time for all Nyxus feature classes versus ROI size for various ROI counts per image when calculated by CPU (grey) and GPU (blue) enabled infrastructure. Figure \ref{figPERF:SCALAB} is assessed across diverse synthetic image mask sizes and counts to enable the user to understand its performance regardless of underlying mask and image size or ROI count. In general, processing time scales linearly when the average ROI size is 500 pixels and larger, regardless of ROI count. As expected on CPU, computation time increases as the size of the ROI increases or the total number of ROIs increases. Nyxus GPU performance was not uniformly faster than CPU performance where there were a large number of small regions, with profiling showed this was due to bottlenecks in transferring data to GPU. This shows that using Nyxus for applications such as whole slide imaging should avoid GPU feature extraction (unless ROIs are larger than $\sim 5,000$ px), since whole slide images can contain more than 100,000 nuclear annotations and would hit the GPU transfer bottlenecks observed here. However, for common cell profiling (hundreds to thousands of $1,000$ pixel ROIs per image) and radiometric profiling, use of GPUs would provide performance gains (Figure \ref{figPERF:SCALAB}).  Figure \ref{figPERF:GPUNONGPU} highlights this further by showing the compute cost structure of GPU accelerated features on an image of 100x 100,000-pixel ROIs on a GPU-disabled and GPU-enabled node. Figure \ref{figPERF:GPUNONGPU} also shows that for these types of images, (low counts of large regions) GPU acceleration can improve execution time by more than 3x. Figure \ref{figPERF:GPUNONGPU} shows the compute cost structure of featurizing an image of 100x 100,000-pixel ROIs on a GPU-disabled and GPU-enabled p5 node. Similar trends to the above conclusions are seen in out-of-core processing when node memory was limited by multiples of the region size.

These results demonstrate the efficiency of Nyxus in measuring image features across two domain datasets as compared to other libraries and across diverse hardware. Nyxus contains the broadest feature set across all tested libraries, and often outperforms them while measuring a more exhaustive feature set. Thus, Nyxus is a significant advancement in image feature extraction and well designed to scale to large biomedical image datasets.

\subsection{Accuracy and Reproducibility}

A common issue in quantitative image analysis is the lack of image feature validation and reproducibility.\cite{jha_etal_2021} Part of this challenge lies in the lack of consensus-based guidelines and terminology when extracting measurements from images (e.g. calculating image-based biomarkers). Efforts such as the Image Biomarker Standardization Initiative (IBSI)\cite{Zwanenburg_etal_2020_IBSI} and the 4DN-BINA-OME-QUAREP (NBO-Q) specification\cite{hammer_etal_2021} are community driven efforts to develop measurement standards, but lack cross-community adoption. To address feature validation and reproducibility, Nyxus has an IBSI profile that calculates radiometric image features that comply with IBSI standards. There is currently no established standard with accompanying dataset for the cell profiling community and thus no equivalent profile has been implemented in Nyxus. However, Nyxus makes feature hyperparameters available to users and allows them to be specified in profiles. Thus, the Nyxus architecture has been designed to adapt to new standards as they become available, so when a cell profiling standard is established the community can trivially integrate it into Nyxus. 


\begin{figure}[H]
    \captionsetup[subfigure]{labelformat=empty}

	\begin{subfigure}[b]{\wid\textwidth} 
    	\caption{\rotatebox{0}{\fontselector Masked p3 r19 class 5 c0 slide}}
    	\vspace{0.5px} 
    	\includegraphics[height=2.5cm]{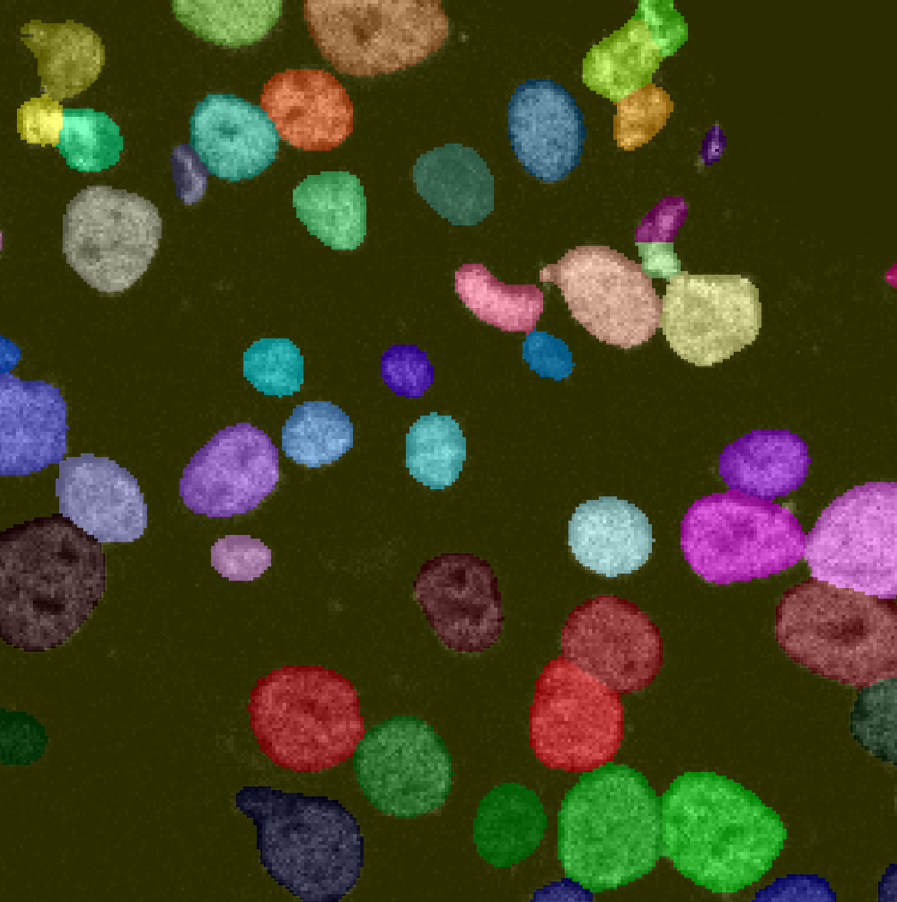}
	\end{subfigure} 
	\hspace{2cm} 
	\input{grafix/autoGeneratedFeatureHeatmaps_IBSI/heatmap_figure_code.tex}

	\caption{Heatmap of fully IBSI-compliant feature extraction results by Nyxus correlated with competition software calculated on a segmented slide hosting 32 ROIs. Color coding: dark blue indicates high correlation, red/orange indicate anti-correlation, and tan/teal represents little/no correlation. } 

	\label{fig:HEATMAP_IBSI} 

\end{figure} 

Features calculated using Nyxus IBSI profile were compared to other libraries, including libraries that comply with IBSI standards. Figure \ref{fig:HEATMAP_IBSI} shows the Pearson correlation of Nyxus feature measurements when the IBSI profile has been selected, where features were calculated on a selected image from the TissueNet 1.0 dataset. Nyxus generally had high correlation of intensity and shape feature class groups relative to Matlab and PyRadiomics, which are libraries also in compliance to IBSI. Of note is a consistently lower correlation between Nyxus and features calculated by RadiomicsJ and CellProfiler. For CellProfiler, a lack of correlation with other feature extraction Methods has been previously reported \cite{bajcsy_etal_2021} and was therefore expected. However, RadiomicsJ was unexpected since it is known to align with the IBSI standard. A preliminary investigation of the low correlation between Nyxus and RadiomicsJ could not establish a good explanation for the differences, but high correlation with PyRadiomics (and the IBSI standard values) lead us to conclude that Nyxus is appropriately compliant with the IBSI standard.

The purpose of the Nyxus IBSI profile is to improve reproducibility in image feature measurements regardless of compute time because IBSI compliance incurs performance costs due to algorithmic complexity (assumption of non-sparse zero-based intensity ranges, zone size, zonal distance, etc). Since the intention of Nyxus more broadly is to have exhaustive, reproducible, and scalable feature sets, an additional profile was created to deliver maximum speed and scalability. The default profile makes minor adjustments to feature extraction algorithms' hyperparameters to significantly improve speed with nominal drops in feature correlation to other IBSI compliant libraries (PyRadiomics). Supplemental Figure \ref{fig:HEATMAP_COMPATIBILITY} shows the correlation between features calculated with the Nyxus default profile and the other libraries. 

This data demonstrates Nyxus compliance with IBSI standards according to the high correlation of Nyxus image feature sets to other established IBSI compliant libraries such as Matlab and PyRadiomics.  Further, Nyxus has additional profiles that improve scalability while still maintaining a high correlation with IBSI compliant libraries (i.e. the default profile) as well as an additional profile that is designed for even more speed for big data at the cost of compliance with IBSI standards (i.e. the performance profile).

\subsection{Architecture and Accessibility}

Nyxus is not only designed for scalability and reproducibility, but also for accessibility and interoperability. Nyxus achieves these goals two different ways: i) interoperability through adoption of scalable open file formats and ii) accessibility through integration with multiple tools and platforms. 

Open hardware initiatives are making images captured on diverse imaging equipment more accessible.\cite{oellermann_open_2022, winter_open-source_2024} The Open Microscopy Environment (OME) has been a leader in the development of tools for reading a wide variety of file formats, and has developed a standard image format (OME-TIFF) with a more recent focus on a next generation file format standard (OME-NGFF). Nyxus reads both OME-TIFF and OME-NGFF standard file formats, as well as the Digital Imaging and Communications in Medicine (DICOM) format, and the Neuroimaging Informatics Technology Initiative (NIfTI) format. Support for the OME-NGFF and NIfTI formats is of particular interest because they are designed for big image data, especially big image data stored in the cloud. In addition to reading open file formats, Nyxus also exports feature measurements into open file formats. Nyxus can export data in the human readable comma separated value (csv) file format, as well as columnar binary file formats from the Apache Foundation such as Arrow and Parquet that are more suitable for feature analytics at scale. Nyxus also natively supports python's Pandas dataframe format in-memory for easy interoperability into typical data science workflows. 

\begin{figure}[H]
	
    
    \centering
    \begin{subfigure}[t]{0.85\textwidth}
        \centering
        \includegraphics[width=\linewidth]{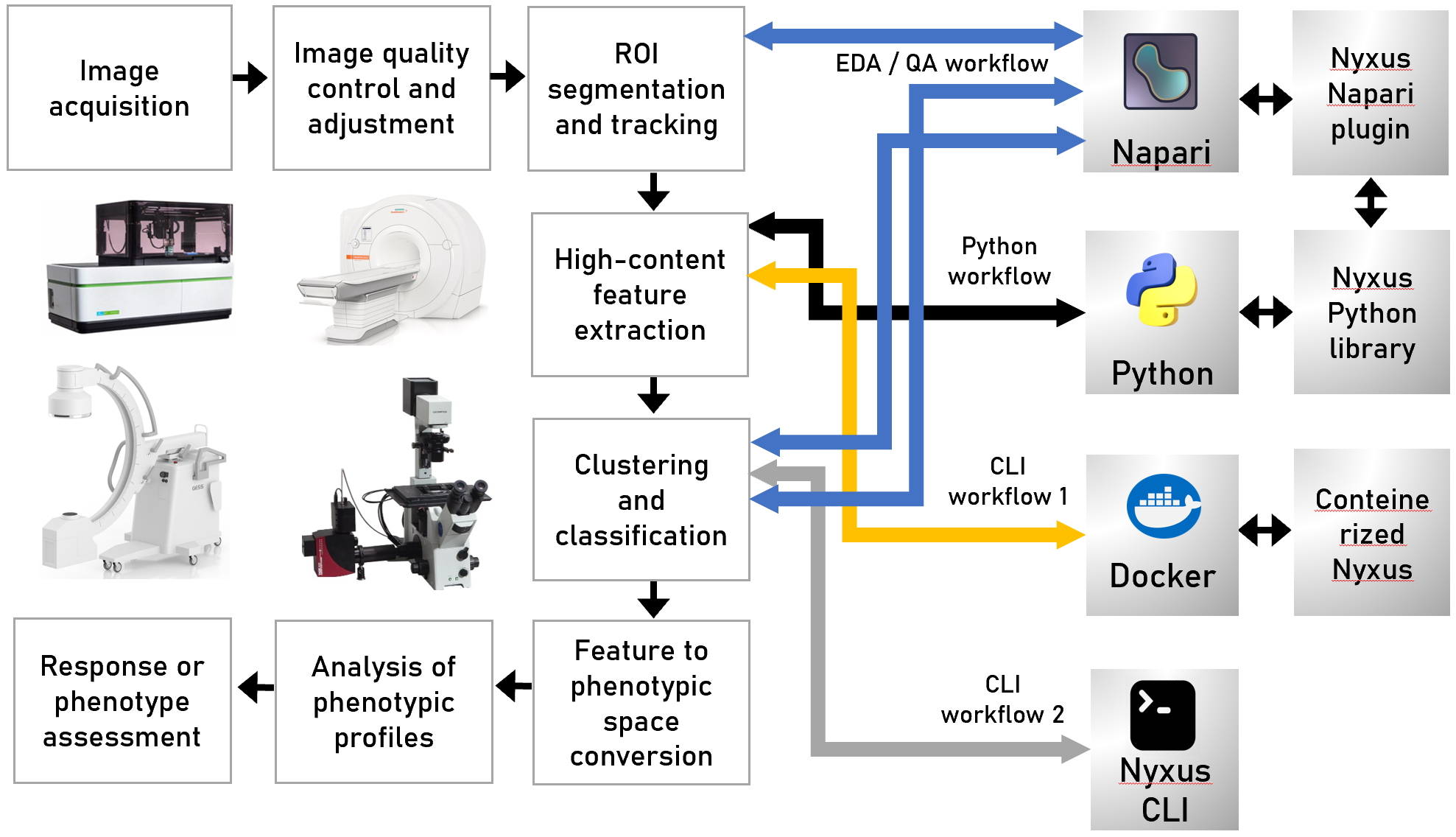}
        \caption{}
        \label{figARCHITECTURE:WORKFLOW}
    \end{subfigure}
	
	\vspace{0.5em}
	
	
    \begin{subfigure}[t]{0.85\textwidth}
        \centering
        \includegraphics[width=\linewidth]{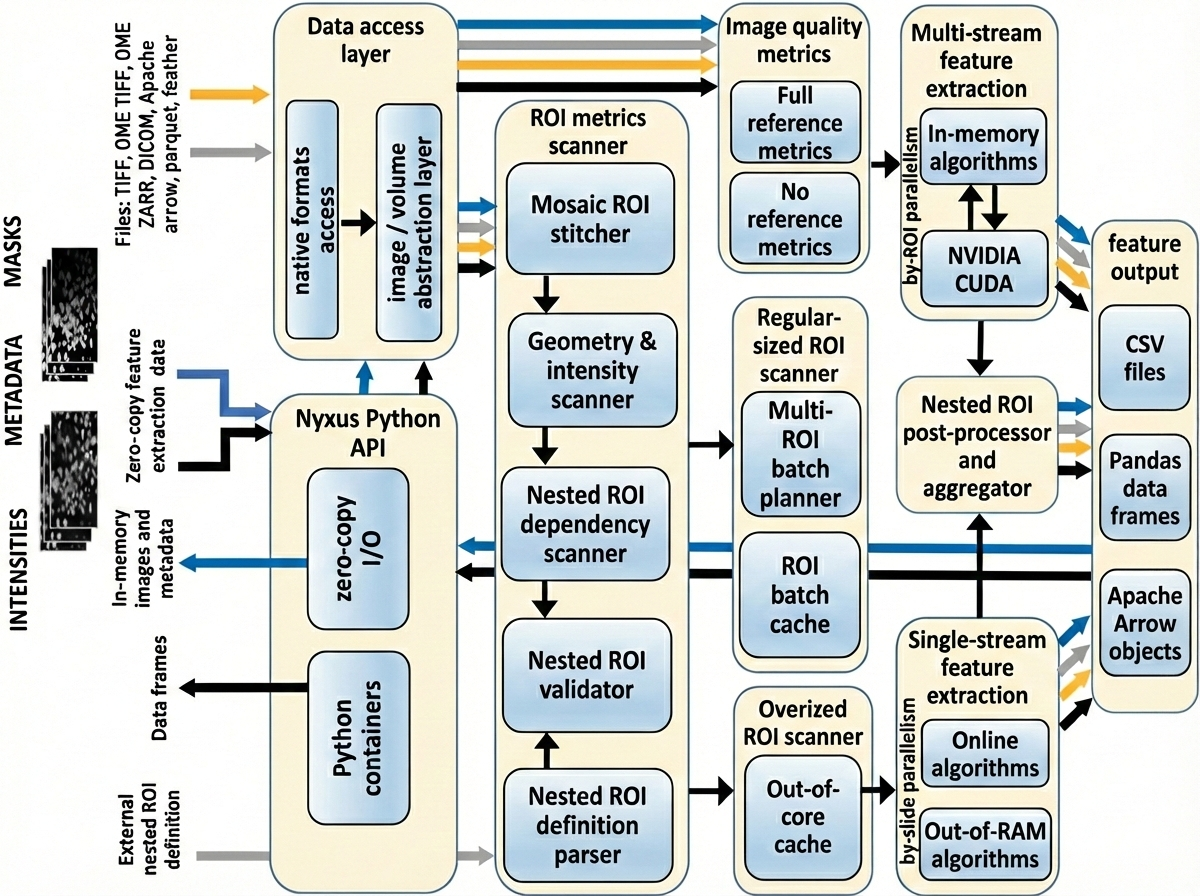}
        \caption{}
        \label{figARCHITECTURE:NYXUSARCH}
    \end{subfigure}
	
	\vspace{0.5em}
	
	
    \begin{subfigure}[t]{0.45\textwidth}
        \centering
        \includegraphics[width=\linewidth]{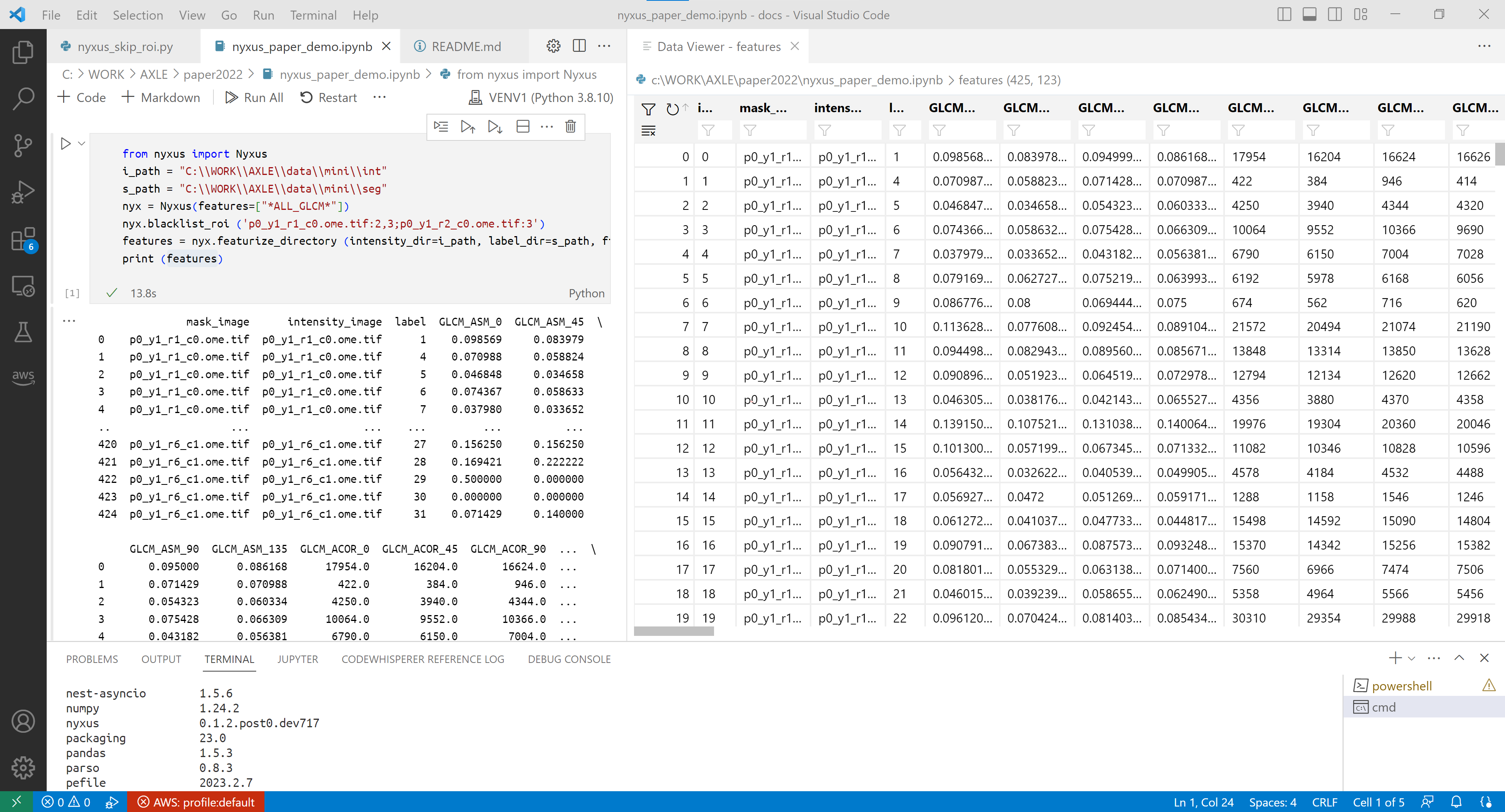}
        \caption{}
        \label{figARCHITECTURE:VSCODE}
    \end{subfigure}
	~
    \begin{subfigure}[t]{0.45\textwidth}
        \centering
        \includegraphics[width=\linewidth]{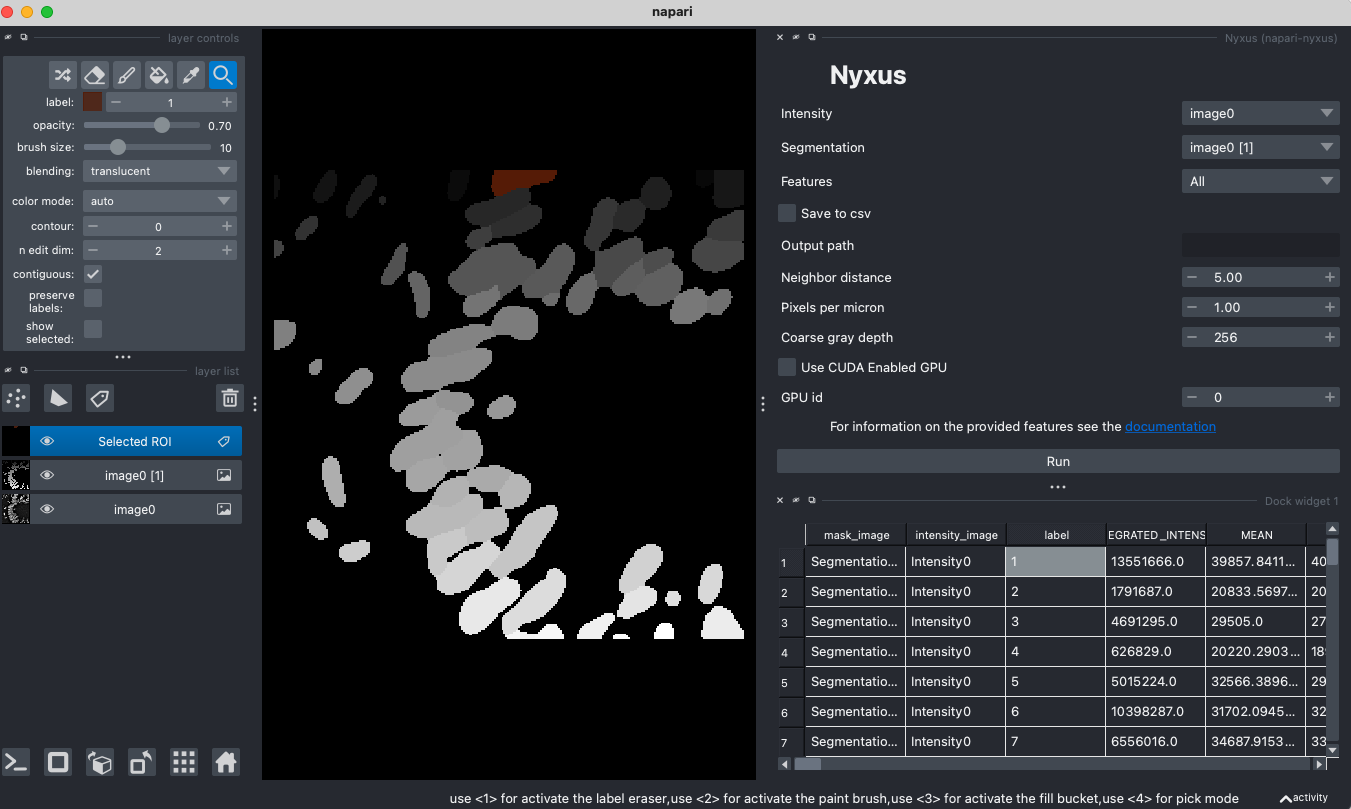}
        \caption{}
        \label{figARCHITECTURE:NAPARI}
    \end{subfigure}
    
    \caption{Nyxus use cases: high-performance pipeline, Python library, and Napari backend. (a) Nyxus serving multiple workflows. (b) Nyxus architecture related to main use cases. (c) Data analysis session with Python. (d) Napari with Nyxus as data analysis backend.}
    
    \label{fig:ARCHITECTURE}
\end{figure}

To make Nyxus accessible to a variety of scientific workflows, many integrations have been implemented. Figure \ref{fig:ARCHITECTURE} provides an overview of Nyxus software, workflows, and integrations. Figure \ref{figARCHITECTURE:WORKFLOW} illustrates the high level workflow for image analysis (acquisition instrument agnostic) and where Nyxus tools fit in a common image analysis workflow. Figure \ref{figARCHITECTURE:NYXUSARCH} presents the software architecture of Nyxus' image feature extraction backend. Colored routes match data exchange and interactions within the image processing workflow. A command-line interface (CLI, shown as a grey line) is available for bulk processing of image data, which is common in High Performance Computing (HPC) applications. The CLI is also provided as a CWL and WIPP tool to make its use easier. For containerized workflows (common in cloud computing), a pre-built Open Container Initiative (OCI) compliant container is available that has been tested with Docker, Podman, or Apptainer runtimes. A Python package as been made available on the Python Package Index (PyPI) for integration into Python tools (purple lane) when customization is desired. Pipeline developers will also find the data structures designed for easy viewing and debugging, as demonstrated by Figure \ref{figARCHITECTURE:VSCODE} which illustrates an interactive session with Nyxus as a Python library. For low and no-code options or to just visualize results, a Napari integration (Figure \ref{figARCHITECTURE:NAPARI}) allows users to extract features and easily visualize metrics extracted from each region of interest.

In addition to the use of open file formats and support for use in a variety of contexts,  Nyxus has an open source codebase that is freely available under the MIT license. The MIT license is highly permissive, allowing use in private, commercial, and public contexts without the need for "copy left" open sourcing of code, permitting use in proprietary or closed source applications. We see this as a critical component to adoption and improvement in reproducibility of scientific and commercial platforms, removing the need for developers and scientists to decide between creating their own algorithms requiring independent validation or open sourcing an entire project to comply with a less permissive copy left license. The open source licensing of Nyxus with the MIT license in addition to the adoption of open standards and large number of integrations helps to maximize the accessibility of Nyxus across use cases and applications.

\section{Conclusion}\label{secCONCL}


We have developed the Nyxus feature extraction library that can extract features in a variety of 2D and 3D datasets produced by biomedical imaging equipment for various modalities. Nyxus reduces the processing time of big image datasets with its focus on efficiency while providing an exhaustive set of flexible features that are configurable to serve multiple modalities/communities and perform targeted and untargeted feature extraction. Our performance results indicate that Nyxus matches or surpasses alternative popular libraries in terms of the number of features extracted and the speed of doing so. Nyxus achieves scalability through its focus on an architecture that utilizes optimized c++ code for multiprocessing and GPU support (when available). By providing a tested and open-source software suite, we aim to create a high-performance, mission-tailored reproducible scientific feature extraction library for biomedical images. In addition to Nyxus' focus on scalability, it also aims to permit reproducible image feature extraction with an option to use IBSI compliant algorithms and a pseudo-IBSI compliant default profile that significantly improves speed at a marginal cost to accuracy. Further, Nyxus is made highly available as a command line tool (CWL and WIPP tool), docker container, Python package, and Napari plugin to support a broad range of users and applications, with a highly permissive open source license for both public and private use.

The need for Nyxus is observed in the rapid advancement in scientific and clinical imaging technologies, marked by increasing resolution, bit-depth, and automation which have led to terabyte and petabyte scale datasets. Nyxus represents a state of the art advancement in image feature extraction across modalities with significant increases in performance to meet the ever increasing size of biomedical image datasets. The team is constantly adding new features to have a more comprehensive set of miscellaneous and volumetric features that will be released during 2026. Future work will focus on inclusion of additional features, more feature parameterization, time series feature extraction, skeleton features, and additional support for other programming languages. The team is open to collaboration and would encourage the community to reach out to us via Github or email for feature requests or integrations. 


\section{Materials and methods}\label{secMATERIALS_AND_METHODS}

\paragraph{\sffamily Datasets}

\begin{figure}[h!]

     \centering
     
    \begin{subfigure}[t]{\textwidth}
        \centering
        \raisebox{-\height}{\includegraphics[width=2cm,height=2cm] {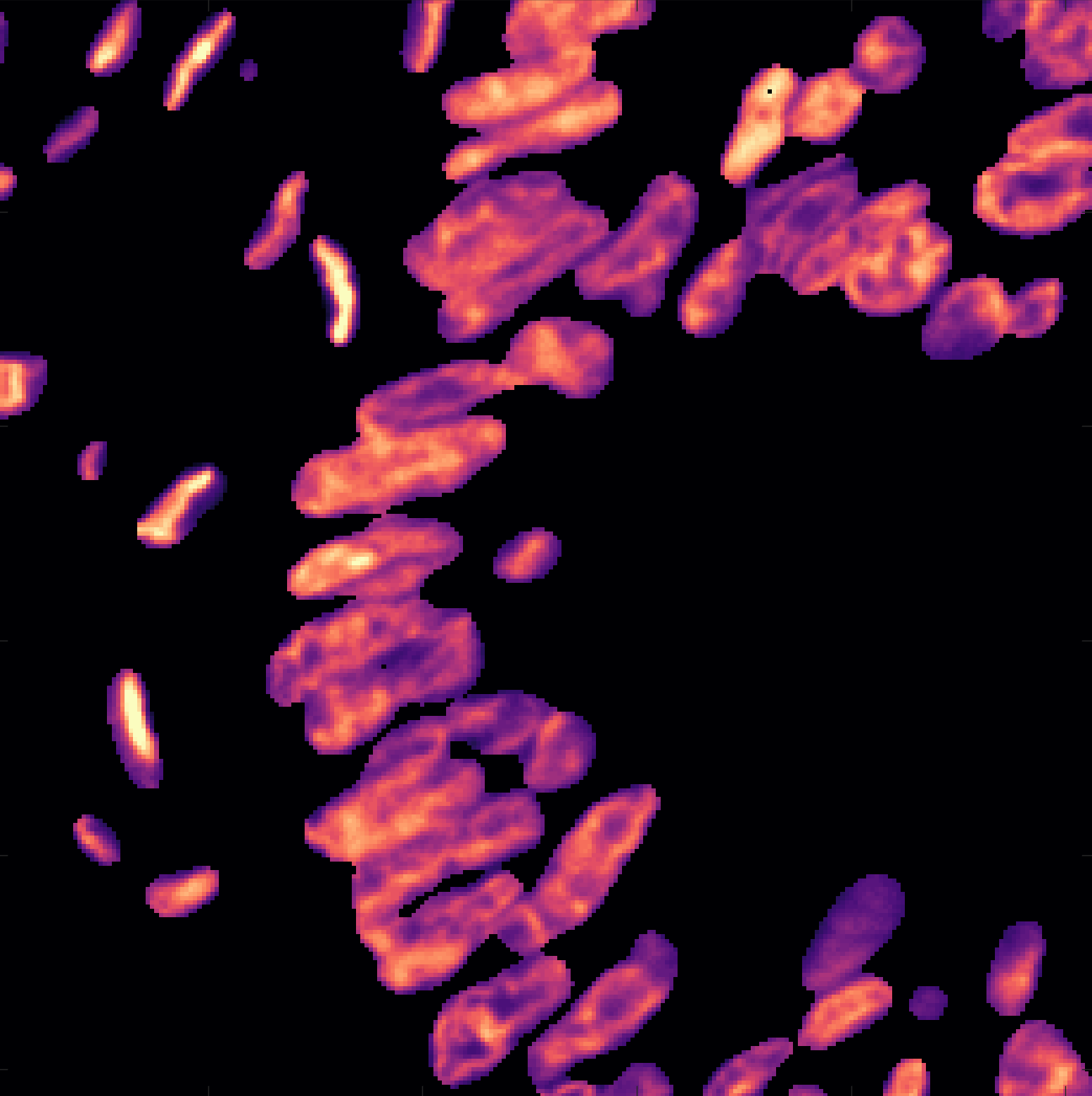}}
        \raisebox{-\height}{\includegraphics[width=2cm,height=2cm] {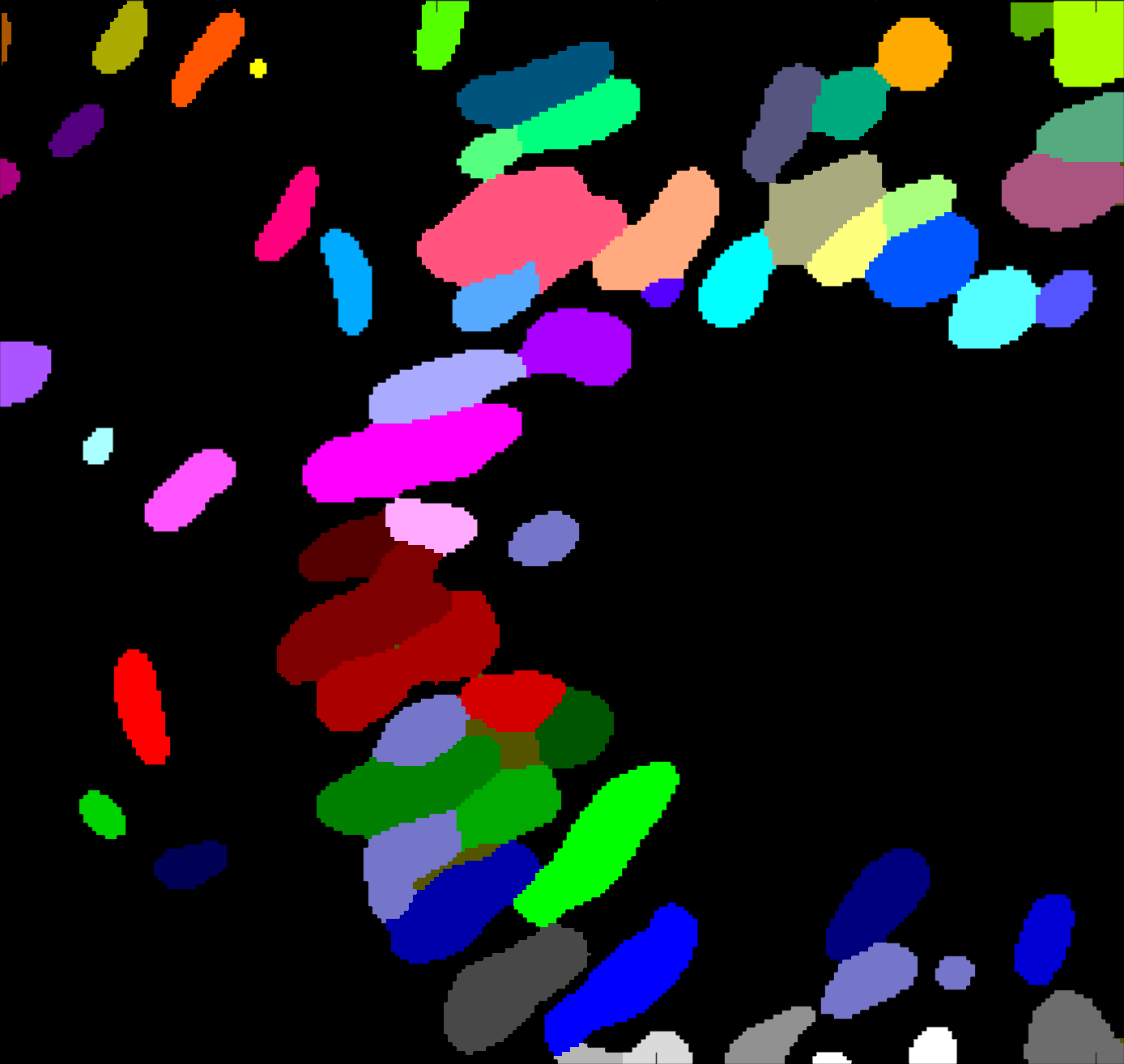}}
        \raisebox{-\height}{\includegraphics[width=2cm,height=2cm] {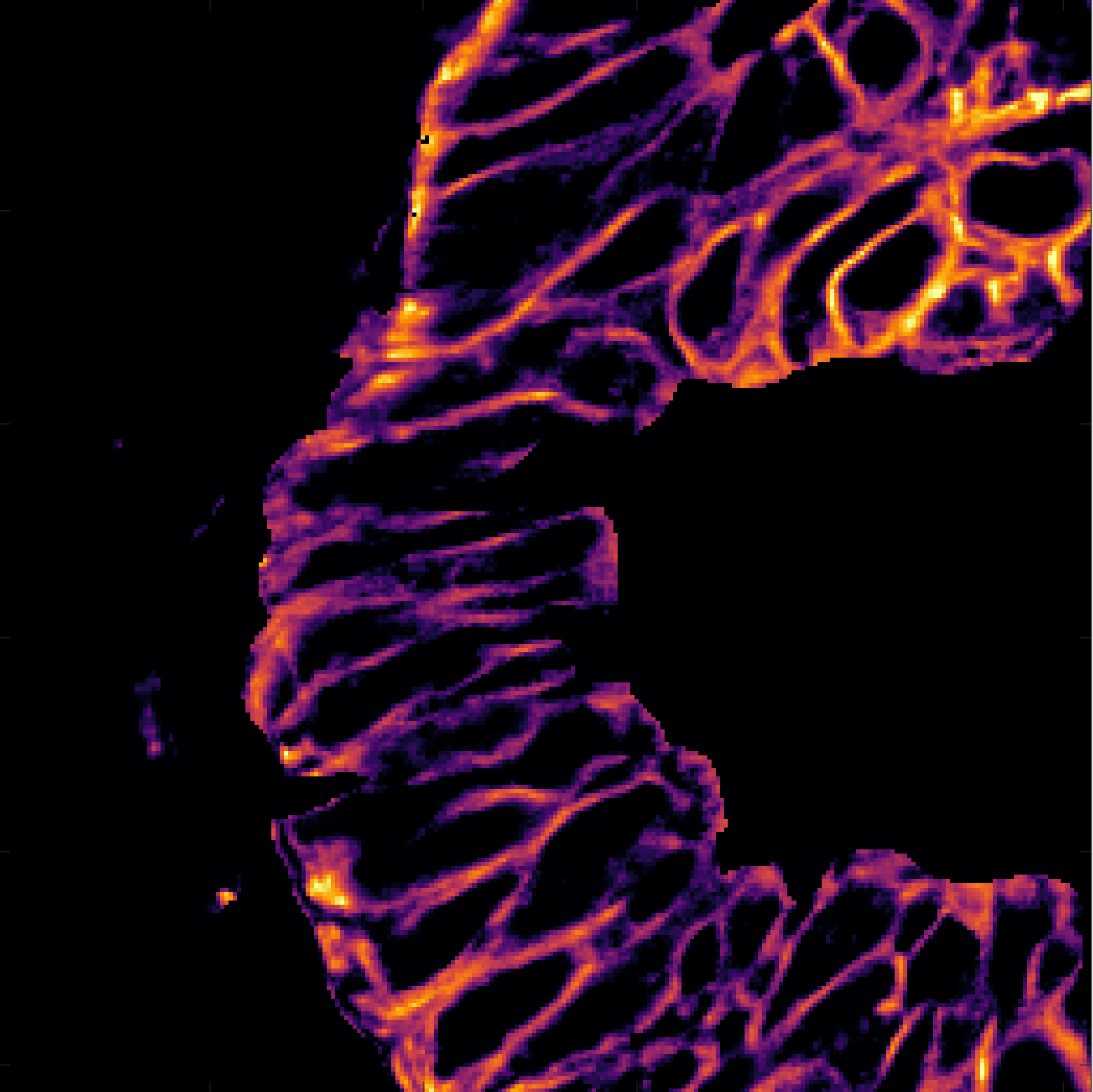}}
        \raisebox{-\height}{\includegraphics[width=2cm,height=2cm] {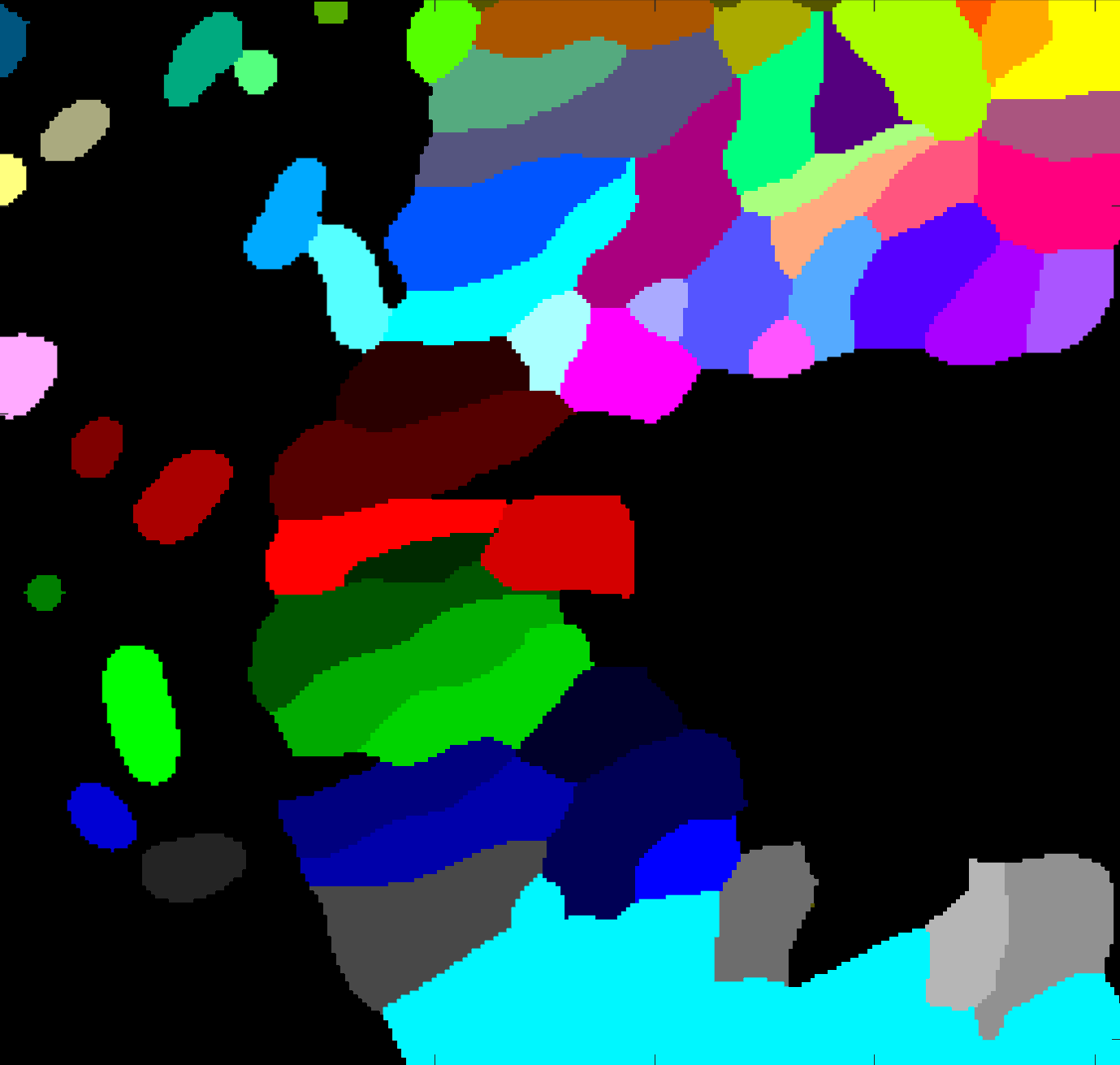}}
        \caption{}
        \label{figDATA:MICROSCOPY}
    \end{subfigure}

    \begin{subfigure}[t]{\textwidth}
        \centering
        \raisebox{-\height}{\includegraphics[width=2cm,height=2cm] {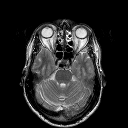}}
        \raisebox{-\height}{\includegraphics[width=2cm,height=2cm] {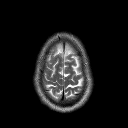}}
        \raisebox{-\height}{\includegraphics[width=2cm,height=2cm] {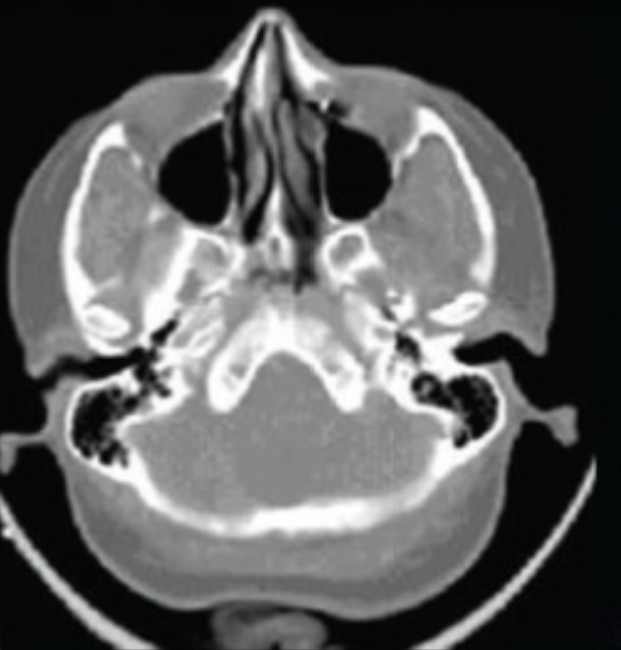}}
        \raisebox{-\height}{\includegraphics[width=2cm,height=2cm] {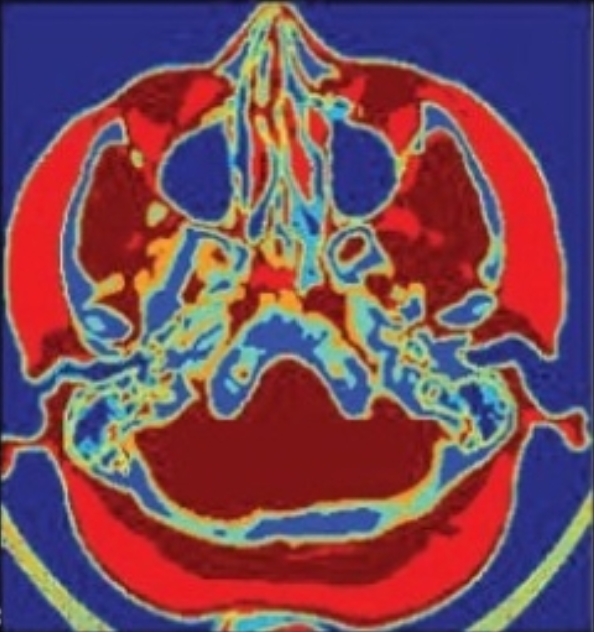}}
        \caption{} 
        \label{figDATA:RADIOMICS}
    \end{subfigure}
    
    \begin{subfigure}[t]{\textwidth}
        \centering
        \raisebox{-\height}{\includegraphics[width=2cm,height=2cm] {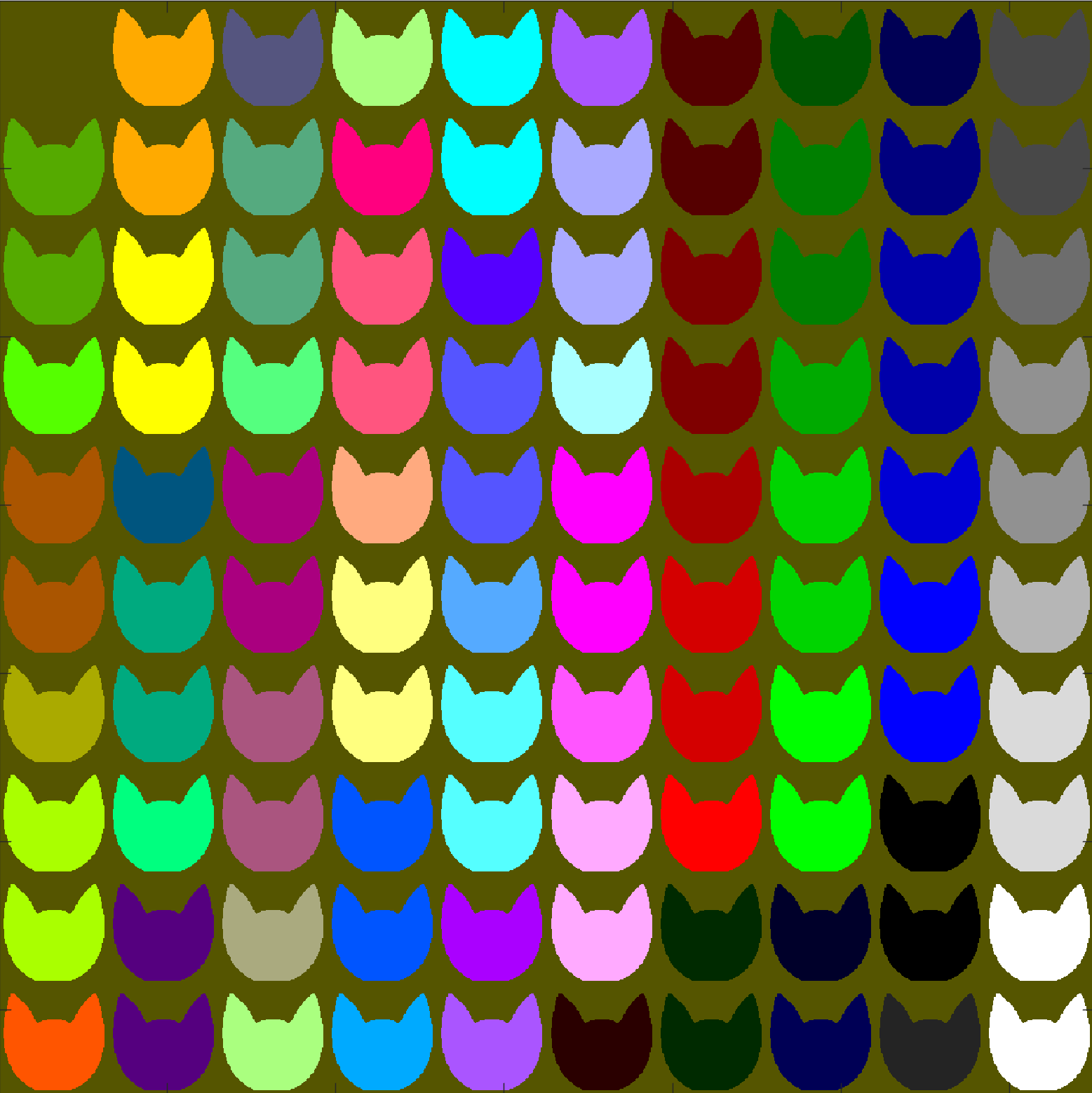}}
        \raisebox{-\height}{\includegraphics[width=2cm,height=2cm] {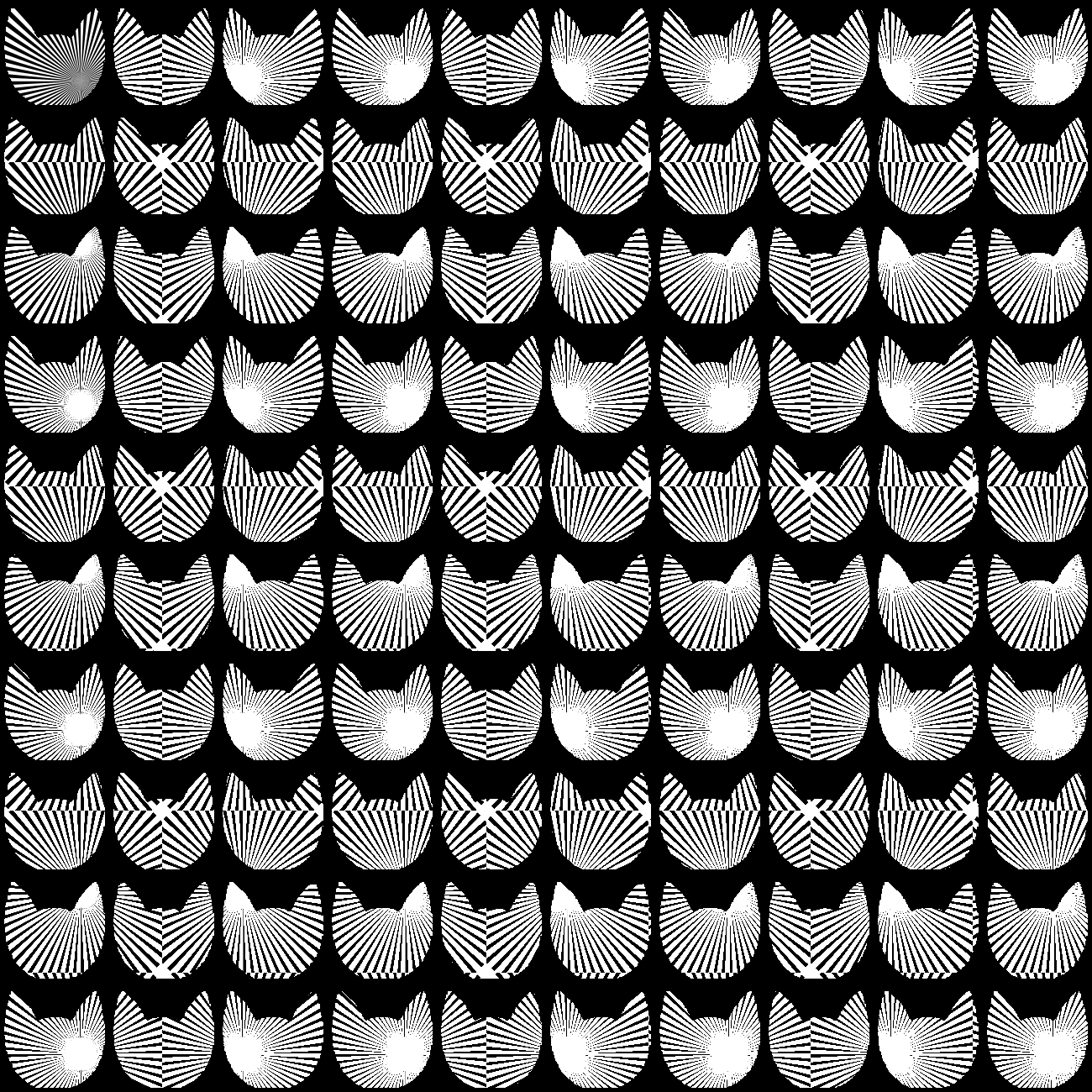}}
        \raisebox{-\height}{\includegraphics[width=2cm,height=2cm] {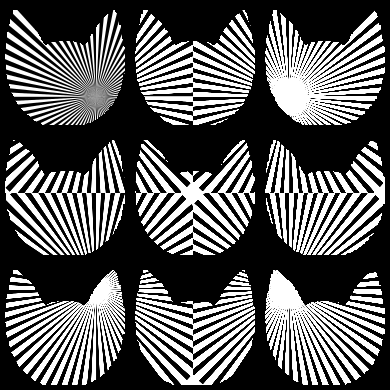}}
        \raisebox{-\height}{\includegraphics[width=2cm,height=2cm] {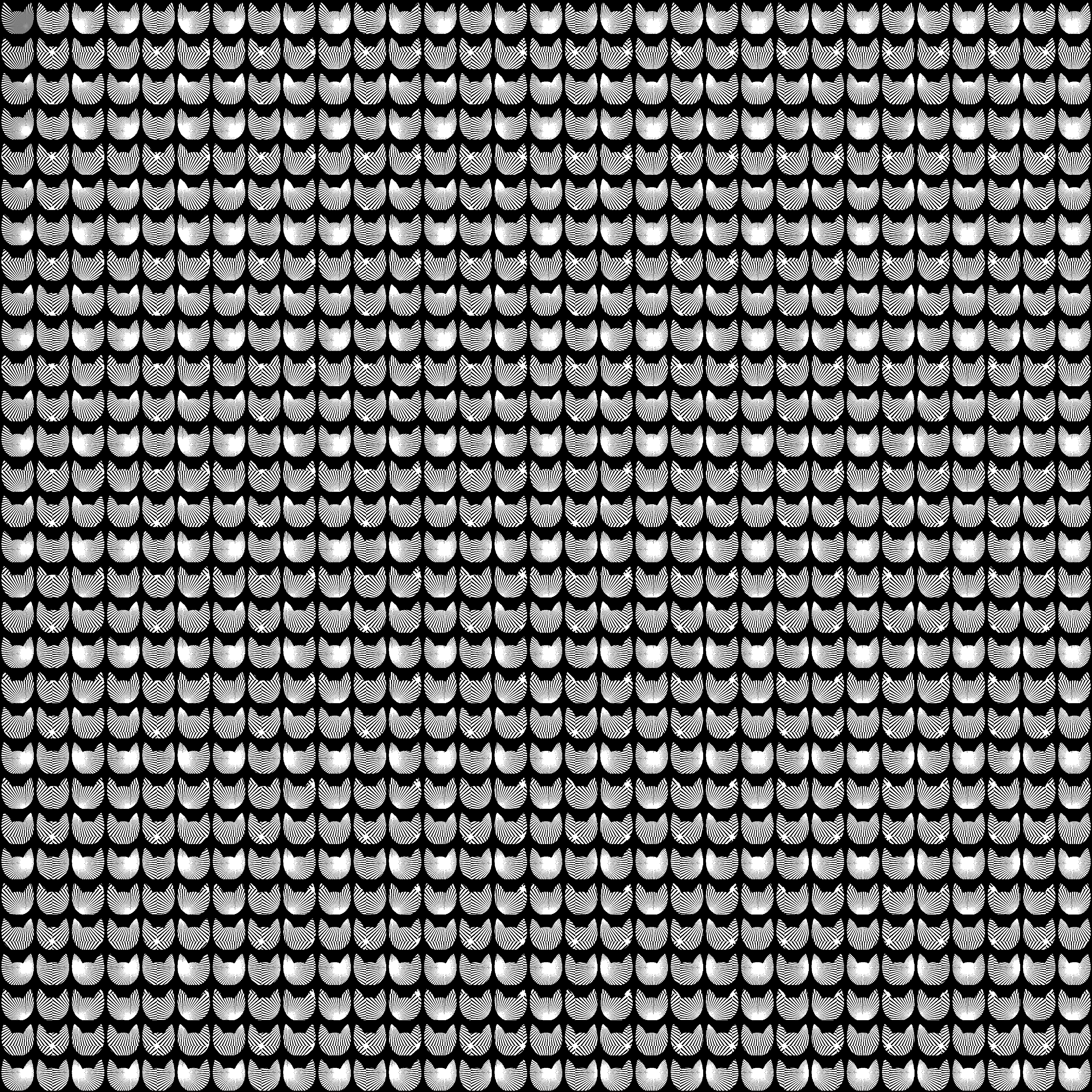}}
        \caption{}
        \label{figDATA:SYNTHETIC}
    \end{subfigure}

    \caption{Representative datasets analyzed. (a) Representative nucleus and cell border images and labeled masks from the TissueNet 1.0 dataset. (b) Representative CT image and labeled masks and MRI image and labeled mask from the Medical Image Decathalon dataset. (c) Representative synthetic data images of labeled masks, and 9, 100, and 1000 binary ROIs overlayed on intensity images.}
    \label{fig:SAMPLEDATA}
\end{figure}

Nyxus was developed to span both the radiomic and cell profiling communities. To show its ability to work across these, two benchmark datasets were analyzed. First, was the TissueNet 1.0 dataset\cite{Greenwald_Miller_Moen_etal_2022} consisting of 3200 fluorescent cell images spanning 6 tissue types.  TissueNet was chosen for the wide range of biological variation (and thus data textures) and because of its breadth of freely available, ground truth, manually segmented images – 2500 total.  Second, was the images from the Medical Segmentation Decathalon.\cite{antonelli_medical_2022}  This dataset was chosen for similar reasons i.e. for its breadth of Magnetic Resonance Imaging (MRI) – 1222 3D and 4D volumes total - and Computed Tomography (CT) images – 1411 volumes total - across 10 different organ/tissue systems leading to a large range of biological variance (and thus data textures) and because of its breadth of freely available ground truth, manually segmented images (2,633 total). The Medical Image Decathalon had the advantage of also having both ground truth segmentations and intensity images across 2D slices as well as total image volumes, enabling testing of Nyxus in both 2D and 3D across diverse image data textures. Figure \ref{figDATA:MICROSCOPY} and \ref{figDATA:RADIOMICS} show representative data from these two benchmark datasets.

In addition to these community standard datasets a synthetic dataset was also generated. 64 intensity images were generated using Siemens optical calibration target\cite{siemens_stern} masked with Arnold cat shape masks. These images scaled in resolution from 100 x 100 to $10^5$ x $10^5$ pixels. In addition to these intensity images, image masks were also generated. The Arnold cat head mask was patterned across binary images matching in resolution to those of the intensity images. The size/resolution of the head was scaled from 100 to $10^5$ and number of masks per image was scaled from 100 to $10^5$. Figure \ref{figDATA:SYNTHETIC} shows representative data from the synthetic datasets. It is hoped that the synthetic dataset can serve as a community benchmark for all non-IBSI features with Nyxus values for these features serving as a reference for the community. All images, the code to produce them, and the Nyxus features extracted from them can be found at.\cite{polus_article_data} 

\paragraph{\sffamily Nyxus architecture}

Nyxus is a free, open-source \cite{polus_nyxus_opensource} software Methodology that has been tested to run on Windows, MacOS (both ARM64 and x86 architectures), and Linux operating systems. Nyxus is written in C++ Version 17 for performance and was bound to Python using Pybind 11 \cite{wenzel_etal_2017_pybind11} for ease of use and adoption by the community. Nyxus has been made available across a large array of options to try to enable users with diverse technical skills to be able to access the tool. Nyxus can be used as a: C++ library, compiled command line tool, Python API (pip installable via PyPI – https://pypi.org/nyxus - or via Conda – https://anaconda.org/conda-forge/nyxus), Open Container Initiative (OCI) compliant container, Web Image Processing Pipeline (WIPP) Plugin, Interoperable Computational Tool (ICT), Common Workflow Language (CWL) command line tool (CLT), and as a Napari Plugin (see Figure \ref{figARCHITECTURE:NAPARI}). Future bindings to Julia and Java are also planned to further expand the community that can leverage Nyxus. Nyxus also comes complete with extensive documentation, https://nyxus.readthedocs.io/en/latest/, that describes not only how to use the tool in the above options, but gives code examples, has a developer guide, and gives in-depth documentation to the formulae used for every feature to improve traceability and reproducibility of analysis with Nyxus. The math background used by Nyxus for all features found in Figure \ref{fig:feature_coverage} can be found at the read the docs link above. 

The main architectural design goals of Nyxus were high separation of concerns and modularity, maximum use of a compiled language backend, and configurable feature measurement algorithms. Nyxus modularity ensures that feature calculation is uncoupled from the infrastructure that prepares the data for input into the feature calculation "store": image preprocessing, ROI separation, etc. – allowing users to simply add features to the feature store and leverage the state-of-the-art without having to become familiar with its inner workings. To further this end, Nyxus creates an internal ROI layout that is represented as a pixel cloud that is sparse, memory-efficient, and extensible in cases of non-trivial ROIs that are too large to be loaded in RAM. Its limit can also be explicitly specified by a user, for example to fit in a memory quota of a shared cloud resource or to take only half the threads and RAM available on the imaging microscope node where images are acquired, allowing the machine to continue to image while running an analysis with Nyxus.  Another design goal was keeping software dependencies to a minimum. Whenever feasible Nyxus was written with its own functions and routines, dramatically reducing the build complexity and size.
Nyxus supports a number of input imaging formats including DICOM, NIfTI, OME-TIFF, and OME-ZARR (ngff), which encapsulate the most popular image formats in each of the Radiomics and Cell Profiling communities. Nyxus Python API can also accept imaging data as an in-memory Numpy array/montage of arrays. On the output side, Nyxus supports writing extracted features to csv, Apache Arrow, Feather, and Parquet as well as an in-memory Pandas array in Python. In both reading and writing next generation formats are used to enable truly scalable and performant image loading, feature extraction, and writing.


\paragraph{\sffamily Feature Control and Tunability}

Feature extraction can be performed in 2 contexts: one where features need to be sensitive to known application-specific use cases e.g. type of tissue, cancer, pathology, deep learning, etc,; the other context does not include any learning and is purely exploratory e.g. non-hypothesis driven profiling. The 2 contexts, targeted and untargeted, require quite different configurations of features based on complex algorithms that involve hyperparameters, for example, bandwidth, number of bins, scan angle and offset of any kind, threshold, etc. Targeted features were shown to be dataset-specific \cite{teh_chin_1988} with feature moment hyperparameters needing to be changed widely in order to capture comprehensive image information. Apart from sensitivity (often controlled with a threshold) a feature hyperparameter often has an impact on the compute time. For example, two hyperparameters used in the calculation of Gabor features with filter banks of 4 (4 single-band angles) and 32 (8x 4 single-band angles) would incur $>5x$ compute time difference. Below is a description of a novel method for feature hyperparameter optimization. A specific implementation is shown for optimization of Gray-Level Co-occurrence Matrix (GLCM) values for maximum predictive accuracy and minimal compute time of an image's class.

Denote feature’s n-component hyperparameter vector as $p=p_1,p_2, ... p_n$ and its compute time as $t(p)$. In general, the same feature’s hyperparameter components belong to different domains

\[
	 \left\{\begin{matrix}
	p_j \in R
	 \\ p_j \in Z
	 \\\{s_1, \: ... \: , s_j\}
	\end{matrix} \right.	
				\tag{1} \label{eq:1}
\]

\noindent where $s_k \in \mathbb{Z}$  or $s_k \in \mathbb{R}$. Consider different values $v$ and $u$ of hyperparameter vector component $j$:  $p_j(v)$ and $p_j(u)$ of the same feature, $v \ne u$. The feature’s algorithm can be organized in a way to ensure a deterministic compute time for integer or real valued hyperparameters
\[
v < u \Rightarrow  t(p|p_j=v) < t(p|p_j=u)
				\tag{2} \label{eq:2}
\]

\noindent for example, due to the fact that  $p_{j(v)}$ involves a larger convolution kernel than $p_{j(u)}$. Or, if $p_j$ is a set, a deterministic algorithm ensures that  

\[
s_1 \subset s_2 \Rightarrow t(p|p_j=s_1) < t(p|p_j=s_2)
				\tag{3} \label{eq:3}
\]

\noindent For example, suppose a set of caliper scan angles $s_2$ is a superset of $s_1$. Extending this principle on a case of 2 hyperparameters, say an m-level integer-valued component $p_j$ and an $n$-item set-valued component $p_k$ whose domains possess properties (\ref{eq:2}) and (\ref{eq:3}), respectively, we can have an ordered value grid 

\[
\begin{matrix}
p_{k_1},p_{j_1} & < & p_{k_1},p_{j_2} & < & ... & < & p_{k_1},p_{j_m} \\
\cap  &  & \cap &  &  &  & \cap \\
p_{k_2},p_{j_1} & < & p_{k_2},p_{j_2} & < & ... & < & p_{k_2},p_{j_m} \\
\cap &  & \cap &  &  &  & \cap \\
... &  & ... &  &  &  & ... \\
\cap &  & \cap &  &  &  & \cap \\
p_{k_n},p_{j_1} & < & p_{k_n},p_{j_2} & < & ... & < & p_{k_n},p_{j_m} \\
\end{matrix}
				\tag{4} \label{eq:4}
\]

\noindent delivering a grid of ordered compute times: 

\[
\begin{matrix}
t(p|p_{k_1},p_{j_1}) & < & t(p|p_{k_1},p_{j_2}) & < & ... & < & t(p|p_{k_1},p_{j_m}) \\
\wedge  &  & \wedge &  &  &  & \wedge \\
t(p|p_{k_2},p_{j_1}) & < & t(p|p_{k_2},p_{j_2}) & < & ... & < & t(p|p_{k_2},p_{j_m}) \\
\wedge &  & \wedge &  &  &  & \wedge \\
... &  & ... &  &  &  & ... \\
\wedge &  & \wedge &  &  &  & \wedge \\
t(p|p_{k_n},p_{j_1}) & < & t(p|p_{k_n},p_{j_2}) & < & ... & < & t(p|p_{k_n},p_{j_m}) \\
\end{matrix}
				\tag{5} \label{eq:5}
\]

\noindent Let’s denote features whose hyperparameter set can and cannot be organized as (\ref{eq:4}-\ref{eq:5}) soft and hard features, respectively. While the compute time of a soft feature is deterministic under its parameter vector, its impact on the feature value variation in different known classes is unknown and is strictly dataset-dependent because each feature value is a certain filter response. The feature algorithm can utilize property (\ref{eq:5}) of a corresponding soft feature to plan a time-optimal feature calculation by choosing a parameter vector delivering a time-accuracy tradeoff.

A common property of any feature extractor is the organization of its functionality as an array of feature calculation algorithms of different computational complexity and degree of freedom in terms of algorithm-specific hyperparameters specified externally. For example, if m is image size, while calculating first order statistics requires $O(m)$ time, the entropy feature calculation involving a histogram of predefined number of intensity bins h requires $O(m,h)$ time; texture features involve two hyperparameters – the intensity scale factor $g$ and the spatial offset $s$ resulting in compute times with 3 degrees of freedom $O(m,g,s)$. Finally, features designed as a bank of filters require 5 hyperparameters to configure the number and characteristics of the filters. 

Due to $card(p) \sim t(p)$, we can plan the feature calculation in a categorization context with maximum accuracy within reasonable compute time. If categorization is not an objective, the other objective, profiling, p can be chosen to deliver a desire granularity of feature values. Unlike $t(p)$, the accuracy is non-deterministic depending on a number of factors including a prediction model type and its parameters, and the training set. We can vaguely interpret the accuracy of predicting classes $a$ and $b$: the larger the margin between feature sets of vectors $F_a$ and $F_b$, the better sets $F_a$ and $F_b$ are linearly separable, the better a model’s accuracy is expected. 

Linear separability of sets has a number of metrics: within and between class variance ratio, Kolmogorov distance, total variation, etc. They can be used to find the optimal $p$ on a small representative calibration subset of examples $X_\textrm{CAL}$ meeting requirements $card(X_{\textrm{CAL}}) \ll card(X)$  and $I(X_{\textrm{CAL}}) \approx I(X)$ where $I(.)$ is the Shannon information gain. The calibration to sample set size ratio can be chosen in a number of ways, particularly, by involving the inference complexity, as discussed in.\cite{oymak_etal_generalization_guarantees_2021,guyon_scaling_law_1997} Generally, finding an optimal feature parameter vector can be expressed as solving a minimax problem:

\[
\min_p \max_{X_{\textrm{CAL}}} L(p)
				\tag{6} \label{eq:6}
\]

\noindent where $L(p)$ is loss at some parameter vector $p$, for example as a ratio of between- and within-class variance \cite{Jankowski_Grabczewski_2006}

\[
\frac {s^2_{\textrm{\;between}}} {s^2_{\textrm{\;within}}}
				\tag{7} \label{eq:7}
\]

\noindent or its class-unbiased form \cite{lomax_2007}

\[
L(p)=\frac { \big( N-K \big) \sum_{i=1}^K n_i {\big ( \overline {f(p)_i} - \overline {f(p)} \big) }^2} {\big(K-1\big)\sum_{i=1}^K \sum^{n_i}_{j=1} {\big ( f(p)_{ij} - \overline {f(p)_i} \big )}^2 }	\tag{8} \label{eq:8}
\] 

\noindent where $N=card(X_\textrm{CAL})$, $K$ – number of classes, and $f(p)_i$ – feature value in class $i$.

\begin{figure}[H]
	
    
    \centering
    \begin{subfigure}[t]{0.95\textwidth}
        \centering
        \includegraphics[width=\textwidth]{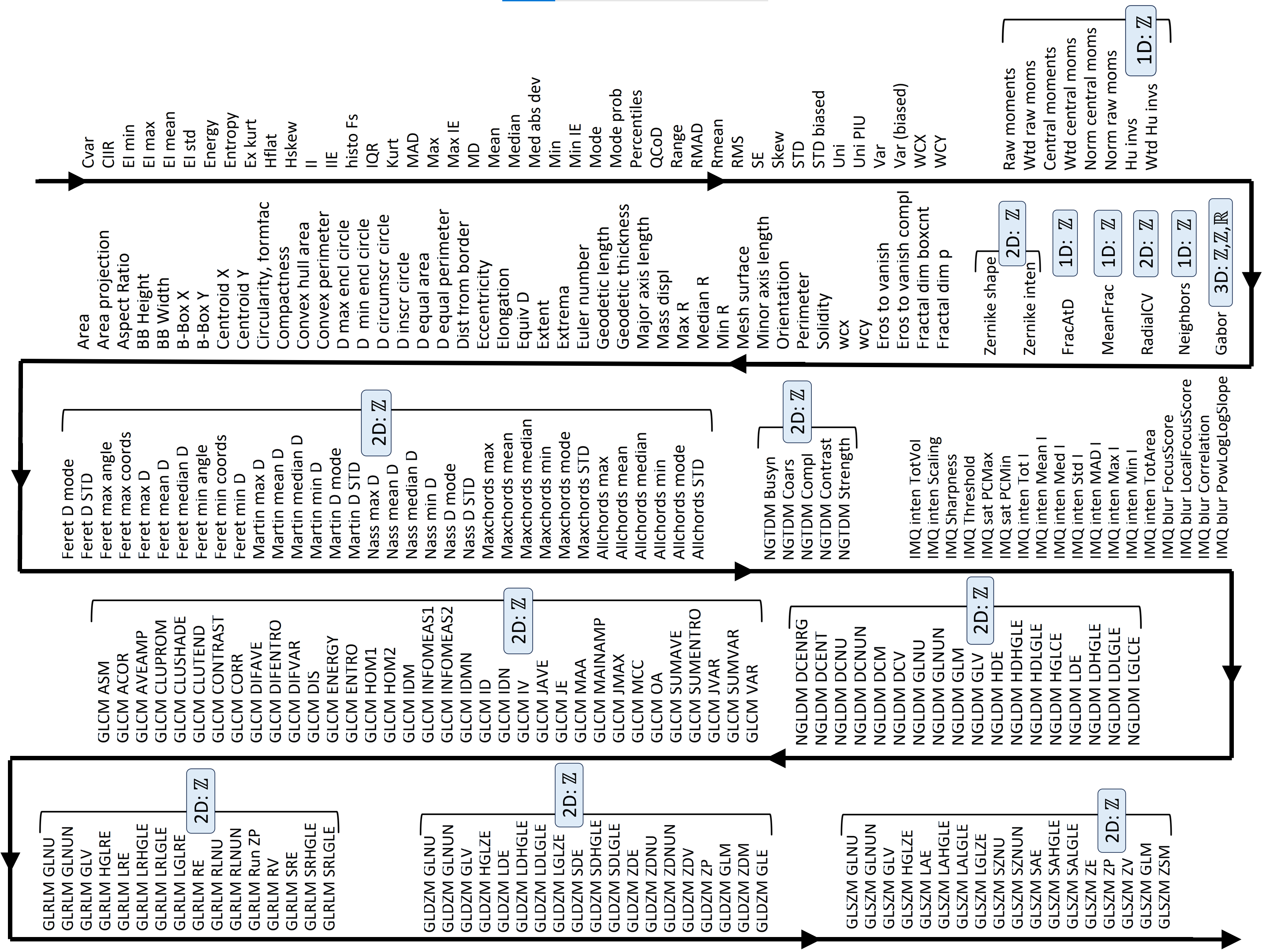}
        \caption{}
        \label{figSOFTFEATURES:TYPICAL}
    \end{subfigure}

	
    \begin{subfigure}[t]{0.63\textwidth}
        \centering
        \includegraphics[width=\textwidth,keepaspectratio]{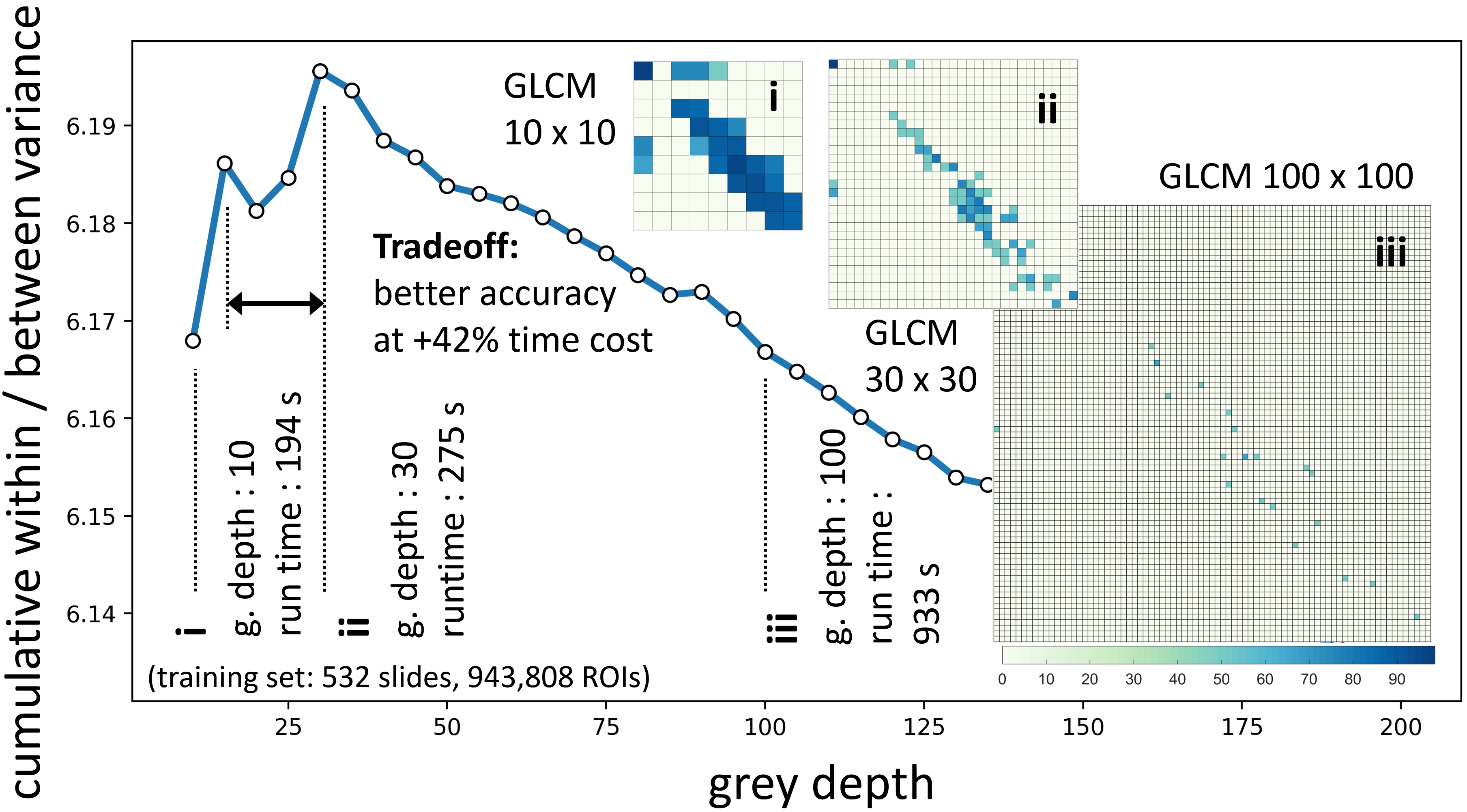}
        \caption{}
        \label{figSOFTFEATURES:GLCM}
    \end{subfigure}
    \hfill
    \begin{subfigure}[t]{0.32\textwidth}
        \centering
        \includegraphics[width=\textwidth,keepaspectratio]{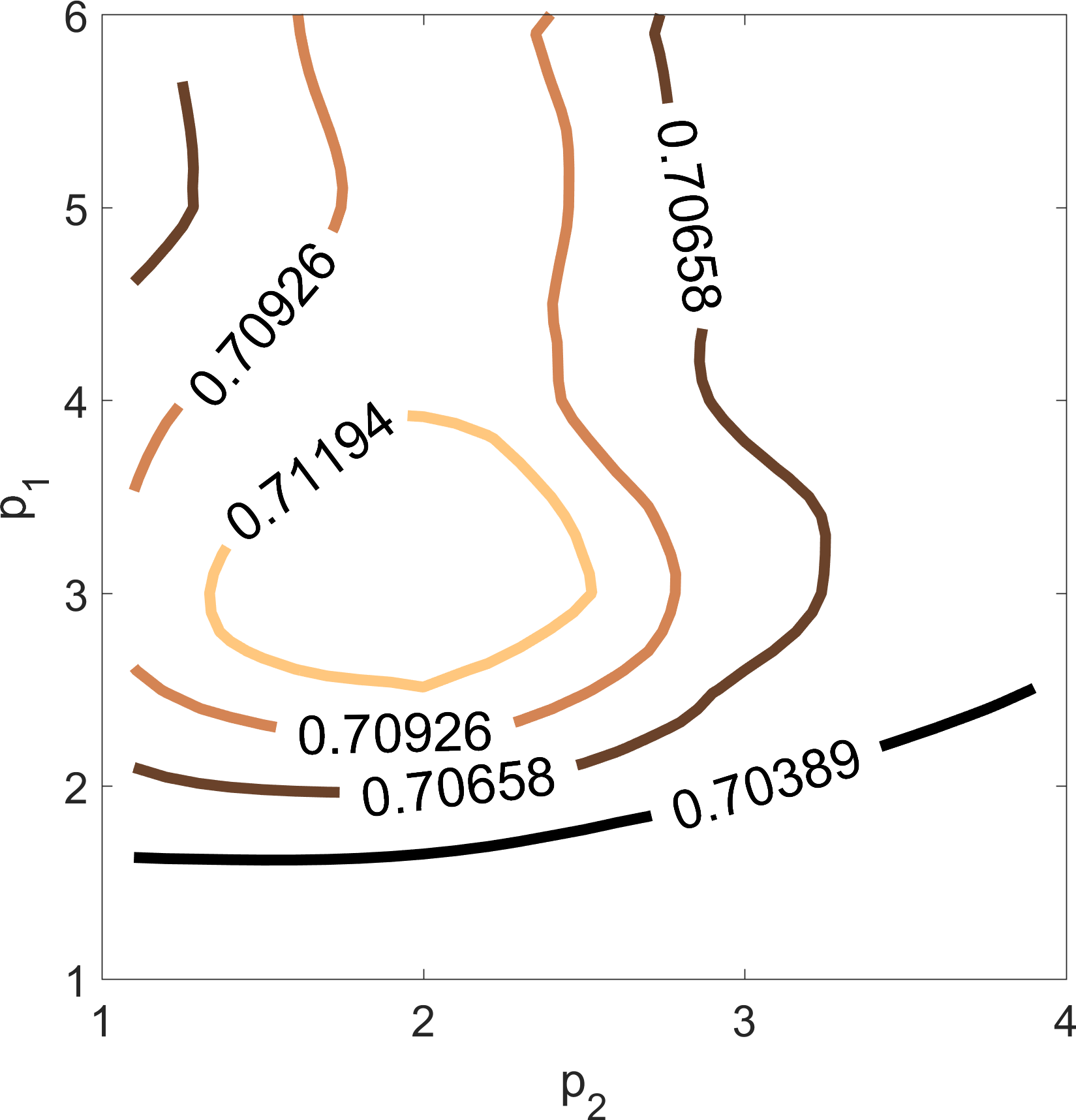}
        \caption{}
        \label{figSOFTFEATURES:CONTOURS}
    \end{subfigure}
    
    \caption{Hard and soft features, and their tuning with Nyxus. 
(a) Typical feature set implementing targeted measurements via soft features. The figure shows all the feature classes that are hard (plain text) and soft (light blue box). 
(b) Tuning a 2D soft feature via varying the grey level co-occurrence matrix size. The figure illustrates tuning the GLCM feature for grey depth and offset on ROIs in 523-images from the Tissuenet dataset. 
Callouts A-C represent a GLCM cooccurrence matrix of varying size to represent computational complexity as more grey depth is added. 
(c) Contours of classification accuracy of the tissue type (yellow – high, purple low) from the Tissuenet test dataset using a 50-state random forest model as a function of feature extraction hyperparameters - grey depth and offset.}
    
    \label{fig:SOFTFEATURES}
\end{figure}

Nyxus supports individual tuning of soft features via its Python API that exposes each soft feature hyperparameter for customizing. Figure \ref{figSOFTFEATURES:TYPICAL} shows all the feature classes that are hard (plain text) and soft (light blue box) in Nyxus. Figure \ref{figSOFTFEATURES:GLCM} illustrates tuning GLCM features \cite{lofstedt_etal_2019} using the above described approach having a 2-dimensional parameter vector on a calibration set of ROIs in 523-images from the Tissuenet dataset. Figure \ref{figSOFTFEATURES:GLCM} shows a comparison of one hyperparameter (GLCM matrix size) that can be tuned, and how pixel specific intensity pairs in the GLCM matrices vary with respect to matrix sizes of 10, 30, and 100 in callouts A-C respectively. High values on the diagonal mean many pixels have the same gray level as their neighbors, indicating a smooth or uniform texture. Callout D illustrates the best accuracy of tissue type classification on the test set of the Tissuenet dataset when using a 50-state random forest model using scikit-learn 1.2.2, at $p_1 = 30$. Here the global maximum of 30 of the inverse loss (\ref{eq:7}) was used. As can be seen in this callout, the maximum classification accuracy occurs well below the grey depth of the original image ($\sim 2^3 - 2^4$ vs $2^8$) and well below the maximum offset (1.5-2.5 vs 4) reducing the overall computation time for this feature class by over an order of magnitude from the defaults in most other Methods while improving accuracy of classification. This example demonstrates the dramatic improvements in computation time and accuracy that can be realized using this approach at model training time for each dataset. Performing the above optimization for each soft feature at model training has been found to decrease total computation time of inference by multiple orders of magnitude while improving or maintaining classification/regression accuracy. 

\paragraph{\sffamily Instrumentation}

The feature extraction libraries referenced in this paper are CellProfiler 4.2.1, Imea 0.3.3, Image Processing Toolbox of MATLAB R2024a, MITK 2023.04, NIST Feature2DJava 1.5.0 WIPP plugin, PyRadiomics 3.0.1, RadiomicsJ 2.1.2,\cite{radiomicsj} WND-CHARM 1.60, and Nyxus 0.8.1. All the benchmarks were taken on a 8-core Amazon EC2 P5 node equipped with Ubuntu 20.04 operating system, AMD EPYC 7R32 CPU, 32 Gb of physical memory, NVIDIA A10G graphical card and CUDA framework 11.6. 
In all the experiments, the official CellProfiler 4.2.1 Docker image, Imea 0.3.3 python 3.7 library, MATLAB R2024a’s Image processing Toolkit for Linux, Phenotyping Tools executable of MITK 2023.04 for Linux, Docker image of NIST Feature2DJava 1.5.0 WIPP plugin, Java v.17 build of PyRadiomics 3.0.1, and Linux executables built from source code of WND-CHARM 1.60 and Nyxus 0.8.1 were used.

\bibliography{sn-bibliography}
\clearpage
\section{Supplemental}\label{secSUPPLEM}

\setcounter{figure}{0}
\renewcommand{\thefigure}{S\arabic{figure}}
\setcounter{table}{0}
\renewcommand{\thetable}{S\arabic{table}}

\begin{figure}[H]
	\captionsetup[subfigure]{labelformat=empty}
	\begin{subfigure}[b]{\wid\textwidth} 
    	\caption{\rotatebox{0}{\fontselector Masked p3 r19 class 5 c0 slide}}
    	\vspace{0.5px} 
    	\includegraphics[height=2.5cm]{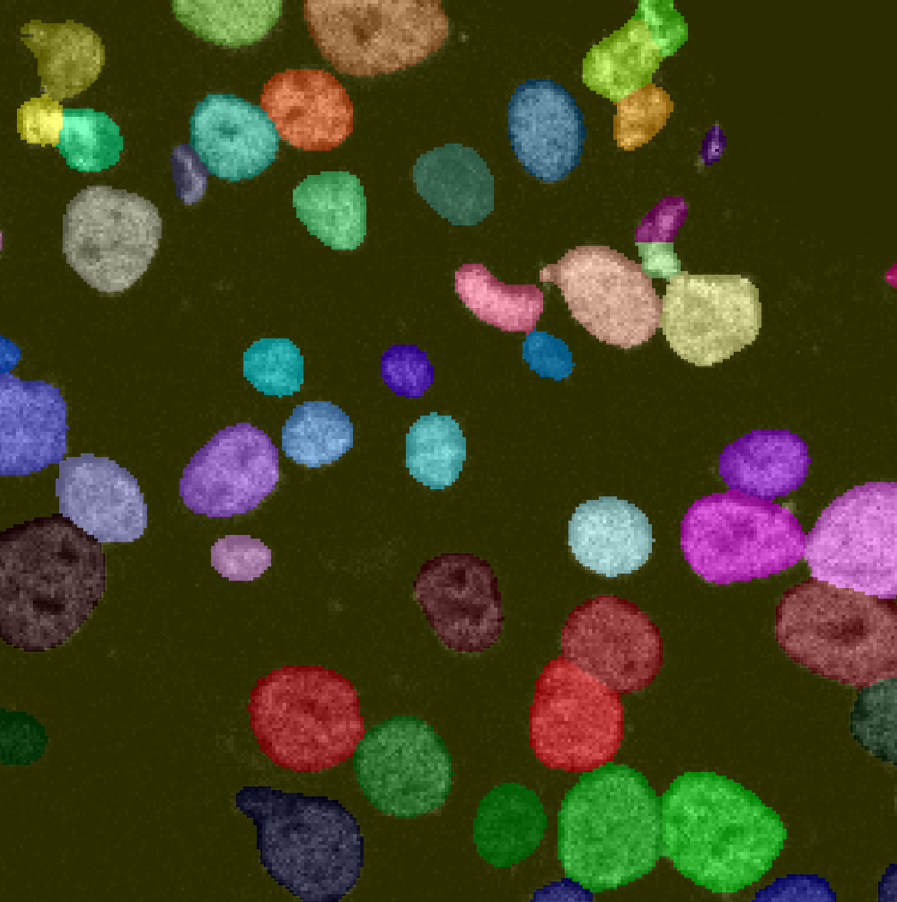}
	\end{subfigure} 
	\hspace{2cm} 
	\input{grafix/autoGeneratedFeatureHeatmaps_NOIBSI/heatmap_figure_code.tex}

	\caption{Heatmap of feature extraction results by Nyxus default profile correlated with competition software calculated on a segmented slide hosting 32 ROIs. Color coding: dark blue indicates high correlation, red/orange indicate anti-correlation, and tan/teal represents little/no correlation.} 

	\label{fig:HEATMAP_COMPATIBILITY} 

\end{figure} 

\renewcommand{\tabcolsep}{3pt}

\clearpage

\begin{table} [!hbt]
\centering
\caption {2-dimensional intensity features}
\label {table:INTEN}
\begin{tiny}
\begin{tabular*}{\textwidth}{@{\extracolsep{\fill}} l  c  c  c  c  c  c  c  c  c  }
\toprule

 Feature & CP &  IM &  MT &  MI &  NF &  PR &  RJ & WC & NY \\
 \midrule
1 C of var & -- & -- & -- & 1 & -- & -- & 1 & -- & -- \\
2 Covered img Inten Range & -- & -- & -- & 1 & -- & -- & -- & -- & -- \\
3 Edge integrated intensity & 1 & -- & -- & -- & -- & -- & -- & 1 & 1 \\
4 Edge intensity max & 1 & -- & -- & -- & -- & -- & -- & -- & 1 \\
5 Edge intensity min & 1 & -- & -- & -- & -- & -- & -- & -- & 1 \\
6 Edge intensity mean & 1 & -- & -- & -- & -- & -- & -- & -- & 1 \\
7 Edge intensity std & 1 & -- & -- & -- & -- & -- & -- & -- & 1 \\
8 Energy & -- & -- & -- & 1 & 1 & 1 & 1 & 1 & 1 \\
9 Entropy & -- & -- & 1 & 1 & 1 & 1 & -- & 1 & 1 \\
10 Excess kurtosis & -- & -- & -- & 1 & -- & -- & -- & -- & -- \\
11 Hyperflatness & -- & -- & -- & -- & 1 & -- & -- & -- & 1 \\
12 Hyperskewness & -- & -- & -- & -- & 1 & -- & -- & -- & 1 \\
13 II & 1 & -- & -- & -- & -- & -- & -- & 1 & 1 \\
14 IQR & -- & -- & 1 & 1 & 1 & 1 & 1 & 1 & 1 \\
15 Kurtosis & -- & -- & 1 & 1 & 1 & 1 & 1 & 1 & 1 \\
16 MD & 1 & -- & -- & -- & -- & -- & -- & 1 & 1 \\
17 Max & 1 & -- & 1 & 1 & -- & 1 & 1 & 1 & 1 \\
18 Mean & 1 & -- & 1 & 1 & 1 & 1 & 1 & 1 & 1 \\
19 MAD & 1 & -- & 1 & 1 & -- & 1 & 1 & 1 & 1 \\
20 Median & 1 & -- & 1 & 1 & 1 & 1 & 1 & 1 & 1 \\
21 Median Abs Dev & -- & -- & -- & 1 & -- & -- & 1 & -- & -- \\
22 Min & 1 & -- & 1 & 1 & -- & 1 & 1 & 1 & 1 \\
23 Mode & -- & -- & 1 & 1 & 1 & -- & -- & 1 & 1 \\
24 Mode probability & -- & -- & -- & 1 & -- & -- & -- & -- & -- \\
25 Percentiles & 1 & -- & 1 & 1 & -- & 1 & 1 & 1 & 1 \\
26 Qu C of disp & -- & -- &  -- &  1 & -- & -- &  1 & -- & -- \\
27 Range & -- & -- & 1 & 1 & -- & 1 & 1 & 1 & 1 \\
28 RMAD & -- & -- & -- & 1 & -- & 1 & -- & 1 & 1 \\
29 Rmean & -- & -- & -- & 1 & -- & -- & -- & -- & -- \\
30 RMS & -- & -- & 1 & 1 & -- & 1 & 1 & 1 & 1 \\
31 SE & -- & -- & 1 & -- & -- & -- & 1 & 1 & 1 \\
32 Skewness & -- & -- & 1 & 1 & 1 & 1 & 1 & 1 & 1 \\
33 std & 1 & -- & 1 & 1 & 1 & 1 & 1 & 1 & 1 \\
34 std biased & -- & -- & -- & 1 & -- & -- & -- & -- & -- \\
35 Uniformity & -- & -- & -- & 1 & -- & 1 & -- & -- & 1 \\
36 Uniformity PIU & -- & -- & -- & -- & -- & -- & -- & -- & 1 \\
37 Variance & -- & -- & 1 & 1 & -- & 1 & 1 & 1 & 1 \\
38 Variance (biased) & -- & -- & -- & 1 & -- & -- & -- & -- & -- \\
39 WCX & 1 & -- & 1 & -- & -- & -- & -- & 1 & 1 \\
40 WCY & 1 & -- & 1 & -- & -- & -- & -- & 1 & 1 \\

\bottomrule
Total (40) & 16 & 0 & 18 & 27 & 11 & 17 & 18 & 23 & 31 \\
\end{tabular*}
\end{tiny}
\vspace{1ex}
\end{table}


\clearpage
\begin{landscape}
\begin{table} [!hbt]
\centering
\caption {2-dimensional shape and geometric moment features}
\label {table:SHAPE_MOMS}
\begin{tiny}
\begin{tabular} {l c c c c c c c c c l c c c c c c c c c }

\toprule
Feature & CP &  IM &  MT &  MI & NF &  PR &  RJ & WC & NY &   Feature & CP &  IM &  MT &  MI & NF &  PR &  RJ & WC & NY \\

\midrule
\hspace{2mm}  \bf Shape \rm &  &  &  &  &  &  &  &  &  & \hspace{2mm}  \bf Statistical lengths \rm &  &  &  &  &  &  &  &  &  \\ 
 41 Area & 1 & 1 & 1 & -- & -- & -- & 1 & 1 & 1 &  89 Feret \diameter mode & -- & 1 & -- & -- & -- & -- & -- & -- & 1 \\
 42 Area projection & -- & 1 & -- & -- & -- & -- & -- & -- & -- &  90 Feret \diameter stddev & -- & 1 & -- & -- & -- & -- & -- & -- & 1 \\
 43 Aspect Ratio & -- & -- & -- & -- & -- & -- & -- & 1 & 1 &  91 Feret max angle & -- & -- & 1 & -- & -- & -- & 1 & -- & -- \\
 44 BB Height & -- & -- & 1 & -- & -- & -- & 1 & 1 & 1 &  92 Feret max coordinates & 1 & -- & 1 & -- & -- & -- & -- & -- & 1 \\
 45 BB Width & -- & -- & 1 & -- & -- & -- & 1 & 1 & 1 &  93 Feret max \diameter & 1 & 1 & 1 & -- & -- & -- & -- & -- & 1 \\
 46 Bounding Box X & -- & -- & 1 & -- & -- & -- & -- & 1 & 1 &  94 Feret mean \diameter & -- & 1 & -- & -- & -- & -- & -- & -- & 1 \\
 47 Bounding Box Y & -- & -- & 1 & -- & -- & -- & -- & 1 & 1 &  95 Feret median \diameter & -- & 1 & -- & -- & -- & -- & -- & -- & 1 \\
 48 Centroid X & 1 & -- & 1 & -- & -- & -- & -- & 1 & 1 &  96 Feret min angle & -- & -- & 1 & -- & -- & -- & -- & -- & -- \\
 49 Centroid Y & 1 & -- & 1 & -- & -- & -- & -- & 1 & 1 &  97 Feret min coordinates & -- & -- & 1 & -- & -- & -- & -- & -- & 1 \\
 50 Circularity, formfactor & 1 & -- & 1 & -- & -- & -- & 1 & 1 & 1 &  98 Feret min \diameter & 1 & 1 & 1 & -- & -- & -- & -- & -- & 1 \\
 51 Compactness & 1 & -- & -- & -- & -- & -- & -- & 1 & 1 &  99 Martin max \diameter & -- & 1 & -- & -- & -- & -- & -- & -- & 1 \\
 52 Convex hull area & -- & -- & 1 & -- & -- & -- & -- & 1 & 1 &  100 Martin mean \diameter & -- & 1 & -- & -- & -- & -- & -- & -- & 1 \\
 53 Convex perimeter & -- & 1 & -- & -- & -- & -- & -- & -- & -- &  101 Martin median \diameter & -- & 1 & -- & -- & -- & -- & -- & -- & 1 \\
 54 \diameter max encl circle & -- & 1 & -- & -- & -- & -- & -- & -- & -- &  102 Martin min \diameter & -- & 1 & -- & -- & -- & -- & -- & -- & 1 \\
 55 \diameter min encl circle & -- & 1 & -- & -- & -- & -- & -- & 1 & 1 &  103 Martin \diameter mode & -- & 1 & -- & -- & -- & -- & -- & -- & 1 \\
 56 \diameter circumscr circle & -- & 1 & -- & -- & -- & -- & -- & 1 & 1 &  104 Martin \diameter stddev & -- & 1 & -- & -- & -- & -- & -- & -- & 1 \\
 57 \diameter inscr circle & -- & 1 & -- & -- & -- & -- & -- & 1 & 1 &  105 Nassenstein max \diameter & -- & 1 & -- & -- & -- & -- & -- & -- & 1 \\
 58 \diameter equal area & -- & 1 & -- & -- & -- & -- & -- & 1 & 1 &  106 Nassenstein mean \diameter & -- & 1 & -- & -- & -- & -- & -- & -- & 1 \\
 59 \diameter equal perimeter & -- & 1 & -- & -- & -- & -- & -- & 1 & 1 &  107 Nassenstein median \diameter & -- & 1 & -- & -- & -- & -- & -- & -- & 1 \\
 60 Dist from border & -- & -- & -- & -- & -- & 1 & -- & -- & -- &  108 Nassenstein min \diameter & -- & 1 & -- & -- & -- & -- & -- & -- & 1 \\
 61 Eccentricity & 1 & -- & 1 & -- & -- & -- & -- & 1 & 1 &  109 Nassenstein \diameter mode & -- & 1 & -- & -- & -- & -- & -- & -- & 1 \\
 62 Elongation & -- & -- & -- & -- & -- & -- & 1 & 1 & 1 &  110 Nassenstein \diameter stddev & -- & 1 & -- & -- & -- & -- & -- & -- & 1 \\
 63 Equiv \diameter & -- & -- & 1 & -- & -- & -- & -- & 1 & 1 &  111 Maxchords max & -- & 1 & -- & -- & -- & -- & -- & 1 & 1 \\
 64 Extent & 1 & -- & 1 & -- & -- & -- & -- & 1 & 1 &  112 Maxchords mean & -- & 1 & -- & -- & -- & -- & -- & 1 & 1 \\
 65 Extrema & -- & -- & 1 & -- & -- & -- & -- & -- & 1 &  113 Maxchords median & -- & 1 & -- & -- & -- & -- & -- & 1 & 1 \\
 66 Euler number & 1 & -- & 1 & -- & -- & -- & -- & 1 & 1 &  114 Maxchords min & -- & 1 & -- & -- & -- & -- & -- & 1 & 1 \\
 67 Geodetic length & -- & 1 & -- & -- & -- & -- & -- & 1 & 1 &  115 Maxchords mode & -- & 1 & -- & -- & -- & -- & -- & 1 & 1 \\
 68 Geodetic thickness & -- & 1 & -- & -- & -- & -- & -- & 1 & 1 &  116 Maxchords stddev & -- & 1 & -- & -- & -- & -- & -- & 1 & 1 \\
 69 Major axis length & 1 & 1 & 1 & -- & -- & -- & 1 & 1 & 1 &  117 Allchords max & -- & 1 & -- & -- & -- & -- & -- & 1 & 1 \\
 70 Mass displacement & 1 & -- & -- & -- & -- & -- & -- & 1 & 1 &  118 Allchords mean & -- & 1 & -- & -- & -- & -- & -- & 1 & 1 \\
 71 Max radius & 1 & -- & -- & -- & -- & -- & -- & 1 & 1 &  119 Allchords median & -- & 1 & -- & -- & -- & -- & -- & 1 & 1 \\
 72 Median radius & 1 & -- & -- & -- & -- & -- & -- & 1 & 1 &  120 Allchords min & -- & 1 & -- & -- & -- & -- & -- & 1 & 1 \\
 73 Min radius & -- & -- & -- & -- & -- & -- & -- & 1 & 1 &  121 Allchords mode & -- & 1 & -- & -- & -- & -- & -- & 1 & 1 \\
 74 Mesh surface & -- & -- & -- & -- & -- & -- & -- & 1 & 1 &  122 Allchords stddev & -- & 1 & -- & -- & -- & -- & -- & 1 & 1 \\
 75 Minor axis length & 1 & 1 & 1 & -- & -- & -- & 1 & 1 & 1 & \hspace{2mm}  \bf Shape geometric moments \rm &  &  &  &  &  &  &  &  &  \\
 76 Orientation & 1 & -- & 1 & -- & -- & -- & -- & 1 & 1 &  123 Raw moments & 1 & -- & -- & -- & -- & -- & -- & -- & 1 \\
 77 Perimeter & 1 & -- & 1 & -- & -- & -- & 1 & 1 & 1 &  124 Weighted raw moments & 1 & -- & -- & -- & -- & -- & -- & -- & 1 \\
 78 Solidity & 1 & -- & 1 & -- & -- & -- & -- & 1 & 1 &  125 Central moments & 1 & -- & -- & -- & -- & -- & -- & -- & 1 \\
 79 Zernike shape features & 1 & -- & -- & -- & -- & -- & -- & -- & -- &  126 Weighted central moments & 1 & -- & -- & -- & -- & -- & -- & -- & 1 \\
 80 WCX & 1 & -- & 1 & -- & -- & -- & -- & 1 & 1 &  127 Normalized central moments & 1 & -- & -- & -- & -- & -- & -- & -- & 1 \\
 81 WCY & 1 & -- & 1 & -- & -- & -- & -- & 1 & 1 &  128 Normalized raw moments & 1 & -- & -- & -- & -- & -- & -- & -- & 1 \\
\hspace{2mm}  \bf Mesodescriptors \rm &  &  &  &  &  &  &  &  &  &  129 Hu invariants & 1 & -- & -- & -- & -- & -- & -- & -- & 1 \\
 82 Erosions to vanish & -- & 1 & -- & -- & -- & -- & -- & 1 & 1 &  130 Weighted Hu invariants & 1 & -- & -- & -- & -- & -- & -- & -- & 1 \\
 83 Eros to vanish compl & -- & 1 & -- & -- & -- & -- & -- & 1 & 1 & \hspace{2mm}  \bf Intensity geometric moments \rm &  &  &  &  &  &  &  &  &  \\
\hspace{2mm}  \bf Microdescriptors \rm &  &  &  &  &  &  &  &  &  &  131 intensity raw moments & 1 & -- & -- & -- & -- & -- & -- & -- & 1 \\
 84 Fractal dim boxcounting & -- & 1 & -- & -- & -- & -- & 1 & 1 & 1 &  132 intensity central moments & 1 & -- & -- & -- & -- & -- & -- & -- & 1 \\
 85 Fractal dim perimeter & -- & 1 & -- & -- & -- & -- & -- & 1 & 1 &  133 intensity normalized raw moments & 1 & -- & -- & -- & -- & -- & -- & -- & 1 \\
\hspace{2mm}  \bf Pattern descriptors \rm &  &  &  &  &  &  &  &  &  &  134 intensity normalized central moments & 1 & -- & -- & -- & -- & -- & -- & -- & 1 \\
 86 Hexagonality ave & -- & -- & -- & -- & -- & -- & -- & 1 & 1 &  135 intensity Hu's moments & 1 & -- & -- & -- & -- & -- & -- & -- & 1 \\
 87 Hexagonality STD & -- & -- & -- & -- & -- & -- & -- & 1 & 1 &  136 intensity weighted raw moments & 1 & -- & -- & -- & -- & -- & -- & -- & 1 \\
 88 Polygonality ave & -- & -- & -- & -- & -- & -- & -- & 1 & 1 &  137 intensity weighted central moments & 1 & -- & -- & -- & -- & -- & -- & -- & 1 \\
   &  &  &  &  &  &  &  &  &  &  138 intensity weighted normalized central moments & 1 & -- & -- & -- & -- & -- & -- & -- & 1 \\
   &  &  &  &  &  &  &  &  &  &  139 intensity weighted Hu's moments & 1 & -- & -- & -- & -- & -- & -- & -- & 1 \\

\bottomrule
Total (99) & 39 & 47 & 27 & 0 & 0 & 1 & 10 & 54 & 92 \\
\end{tabular}
\end{tiny}
\end{table}
\end{landscape}


\clearpage
\begin{table}[!hbt]
\centering
\caption {2-dimensional texture features}
\label {table:TEXTURE}
\begin{tiny}
\begin{tabular} { l  c  c  c  c  c  c  c  c  c    l  c  c  c  c  c  c  c  c  c }
 \toprule
  Feature & CP &  IM &  MT &  MI & NF &  PR &  RJ & WC & NY &   Feature & CP &  IM &  MT &  MI & NF &  PR &  RJ & WC & NY \\
\hspace{2mm}  \bf GLCM \rm  &  &   &   &   &   &   &   &  &  &  197 LRE & -- &  -- &  -- &  1 &  -- &  1 &  -- & -- &  1 \\ 
 140 ASM & 1 &  -- &  -- &  -- &  -- &  1 &  1 & 1 & 1 &  198 LRHGLE & -- &  -- &  -- &  1 &  -- &  1 &  -- & -- &  1 \\
 141 ACOR & -- &  -- &  -- &  -- &  -- &  1 &  1 & -- & 1 &  199 LRLGLE & -- &  -- &  -- &  1 &  -- &  1 &  -- & -- &  1 \\
 142 AVEAMP & -- &  -- &  -- &  -- &  1 &  -- &  -- & 1 & 1 &  200 LGLRE & -- &  -- &  -- &  1 &  -- &  1 &  -- & -- &  1 \\
 143 CLUPROM & -- &  -- &  -- &  -- &  -- &  1 &  1 & -- & 1 &  201 RE & -- &  -- &  -- &  1 &  -- &  1 &  -- & -- &  1 \\
 144 CLUSHADE & -- &  -- &  -- &  -- &  -- &  1 &  1 & -- & 1 &  202 RLNU & -- &  -- &  -- &  1 &  -- &  1 &  -- & -- &  1 \\
 145 CLUTEND & -- &  -- &  -- &  -- &  -- &  1 &  1 & -- & 1 &  203 RLNUN & -- &  -- &  -- &  1 &  -- &  1 &  -- & -- &  1 \\
 146 CONTRAST & 1 &  -- &  1 &  -- &  1 &  1 &  1 & 1 & 1 &  204 Run ZP & -- &  -- &  -- &  1 &  -- &  1 &  -- & -- &  1 \\ 
 147 CORR & 1 &  -- &  1 &  -- &  1 &  1 &  1 & 1 & 1 &  205 RV & -- &  -- &  -- &  1 &  -- &  1 &  -- & -- &  1 \\
 148 DIFAVE & -- &  -- &  -- &  -- &  1 &  1 &  1 & -- & 1 &  206 SRE & -- &  -- &  -- &  1 &  -- &  1 &  -- & -- &  1 \\
 149 DIFENTRO & 1 &  -- &  -- &  -- &  1 &  1 &  1 & 1 & 1 &  207 SRHGLE & -- &  -- &  -- &  1 &  -- &  1 &  -- & -- &  1 \\
 150 DIFVAR & 1 &  -- &  -- &  -- &  1 &  1 &  1 & 1 & 1 &  208 SRLGLE & -- &  -- &  -- &  1 &  -- &  1 &  -- & -- &  1 \\
 151 DIS & -- &  -- &  -- &  -- &  -- &  1 &  1 & -- & 1 & \hspace{2mm}  \bf GLDZM \rm  &  &   &   &   &   &   &   &  &   \\
 152 ENERGY & -- &  -- &  1 &  -- &  1 &  1 &  -- & -- & 1 &  209 GLNU & -- &  -- &  -- &  1 &  -- &  -- &  1 & -- &  1 \\
 153 ENTROPY & 1 &  -- &  -- &  -- &  1 &  -- &  -- & 1 & 1 &  210 GLNUN & -- &  -- &  -- &  1 &  -- &  -- &  1 & -- &  1 \\  
 154 HOM1 & -- &  -- &  1 &  -- &  1 &  1 &  -- & -- & 1 &  211 GLV & -- &  -- &  -- &  1 &  -- &  -- &  1 & -- &  1 \\
 155 HOM2 & -- &  -- &  -- &  -- &  -- &  1 &  -- & -- & 1 &  212 HGLZE & -- &  -- &  -- &  1 &  -- &  -- &  1 & -- &  1 \\ 
 156 IDM & 1 &  -- &  -- &  -- &  1 &  1 &  1 & 1 & 1 &  213 LDE & -- &  -- &  -- &  1 &  -- &  -- &  1 & -- &  1 \\
 157 INFOMEAS1 & 1 &  -- &  -- &  -- &  -- &  1 &  1 & 1 & 1 &  214 LDHGLE & -- &  -- &  -- &  1 &  -- &  -- &  1 & -- &  1 \\
 158 INFOMEAS2 & 1 &  -- &  -- &  -- &  -- &  1 &  1 & 1 & 1 &  215 LDLGLE & -- &  -- &  -- &  1 &  -- &  -- &  1 & -- &  1 \\
 159 IDMN & -- &  -- &  -- &  -- &  -- &  1 &  1 & -- & 1 &  216 LGLZE & -- &  -- &  -- &  1 &  -- &  -- &  1 & -- &  1 \\
 160 ID & -- &  -- &  -- &  -- &  -- &  1 &  1 & -- & 1 &  217 SDE & -- &  -- &  -- &  1 &  -- &  -- &  1 & -- &  1 \\
 161 IDN & -- &  -- &  -- &  -- &  -- &  1 &  1 & -- & 1 &  218 SDHGLE & -- &  -- &  -- &  1 &  -- &  -- &  1 & -- &  1 \\
 162 IV & -- &  -- &  -- &  -- &  -- &  1 &  1 & -- & 1 &  219 SDLGLE & -- &  -- &  -- &  1 &  -- &  -- &  1 & -- &  1 \\
 163 JAVE & -- &  -- &  -- &  -- &  -- &  1 &  1 & -- & 1 &  220 ZDE & -- &  -- &  -- &  1 &  -- &  -- &  1 & -- &  1 \\
 164 JE & -- &  -- &  -- &  -- &  -- &  1 &  1 & -- & 1 &  221 ZDNU & -- &  -- &  -- &  1 &  -- &  -- &  1 & -- &  1 \\
 165 MAA & -- &  -- &  -- &  -- &  1 &  -- &  -- & 1 & -- &  222 ZDNUN & -- &  -- &  -- &  1 &  -- &  -- &  1 & -- &  1 \\
 166 MAINAMP & -- &  -- &  -- &  -- &  1 &  -- &  -- & 1 & -- &  223 ZDV & -- &  -- &  -- &  1 &  -- &  -- &  1 & -- &  1 \\
 167 JMAX & -- &  -- &  -- &  -- &  -- &  1 &  1 & -- & 1 &  224 ZP & -- &  -- &  -- &  1 &  -- &  -- &  1 & -- &  1 \\
 168 MCC & -- &  -- &  -- &  -- &  -- &  1 &  -- & -- & -- &  225 GLM & -- &  -- &  -- &  1 &  -- &  -- &  -- & -- &  1 \\
 169 OA & -- &  -- &  -- &  -- &  1 &  -- &  -- & -- & -- &  226 ZDM & -- &  -- &  -- &  1 &  -- &  -- &  -- & -- &  1 \\
 170 SUMAVE & 1 &  -- &  -- &  -- &  1 &  1 &  1 & 1 & 1 &  227 GLE & -- &  -- &  -- &  1 &  -- &  -- &  -- & -- &  -- \\
 171 SUMENT & 1 &  -- &  -- &  -- &  1 &  1 &  1 & 1 & 1 & \hspace{2mm}  \bf GLSZM \rm  &  &   &   &   &   &   &   &  &   \\
 172 JVAR & -- &  -- &  -- &  -- &  -- &  1 &  1 & -- & 1 &  228 GLNU & -- &  -- &  -- &  1 &  -- &  1 &  1 & -- &  1 \\ 
 173 SUMVAR & 1 &  -- &  -- &  -- &  1 &  1 &  1 & 1 & 1 &  229 GLNUN & -- &  -- &  -- &  1 &  -- &  1 &  1 & -- &  1 \\
 174 VARIANCE & 1 &  -- &  -- &  -- &  1 &  -- &  -- & 1 & 1 &  230 GLV & -- &  -- &  -- &  1 &  -- &  1 &  1 & -- &  1 \\
\hspace{2mm}  \bf NGLDM \rm  &  &   &   &   &   &   &   &  &  &  231 HGLZE & -- &  -- &  -- &  1 &  -- &  1 &  1 & -- &  1 \\ 
 175 DCENRG & -- &  -- &  -- &  -- &  -- &  -- &  1 & -- & 1 &  232 LAE & -- &  -- &  -- &  1 &  -- &  1 &  1 & -- &  1 \\
 176 DCENT & -- &  -- &  -- &  -- &  -- &  1 &  1 & -- & 1 &  233 LAHGLE & -- &  -- &  -- &  1 &  -- &  1 &  1 & -- &  1 \\
 177 DCNU & -- &  -- &  -- &  -- &  -- &  1 &  1 & -- & 1 &  234 LALGLE & -- &  -- &  -- &  1 &  -- &  1 &  1 & -- &  1 \\
 178 DCNUN & -- &  -- &  -- &  -- &  -- &  1 &  1 & -- & 1 &  235 LGLZE & -- &  -- &  -- &  1 &  -- &  1 &  1 & -- &  1 \\
 179 DCM & -- &  -- &  -- &  -- &  -- &  -- &  -- & -- & 1 &  236 SZNU & -- &  -- &  -- &  1 &  -- &  1 &  1 & -- &  1 \\
 180 DCV & -- &  -- &  -- &  -- &  -- &  1 &  1 & -- & 1 &  237 SZNUN & -- &  -- &  -- &  1 &  -- &  1 &  1 & -- &  1 \\
 181 GLNU & -- &  -- &  -- &  -- &  -- &  1 &  1 & -- & 1 &  238 SAE & -- &  -- &  -- &  1 &  -- &  1 &  1 & -- &  1 \\
 182 GLNUN & -- &  -- &  -- &  -- &  -- &  -- &  1 & -- & 1 &  239 SAHGLE & -- &  -- &  -- &  1 &  -- &  1 &  1 & -- &  1 \\
 183 GLM & -- &  -- &  -- &  -- &  -- &  -- &  -- & -- & 1 &  240 SALGLE & -- &  -- &  -- &  1 &  -- &  1 &  1 & -- &  1 \\
 184 GLV & -- &  -- &  -- &  -- &  -- &  1 &  1 & -- & 1 &  241 ZE & -- &  -- &  -- &  1 &  -- &  1 &  1 & -- &  1 \\
 185 HDE & -- &  -- &  -- &  -- &  -- &  1 &  1 & -- & 1 &  242 ZP & -- &  -- &  -- &  1 &  -- &  1 &  1 & -- &  1 \\
 186 HDHGLE & -- &  -- &  -- &  -- &  -- &  1 &  1 & -- & 1 &  243 ZV & -- &  -- &  -- &  1 &  -- &  1 &  1 & -- & 1 \\
 187 HDLGLE & -- &  -- &  -- &  -- &  -- &  1 &  1 & -- & 1 &  244 GLM & -- &  -- &  -- &  1 &  -- &  -- &  -- & -- & -- \\
 188 HGLCE & -- &  -- &  -- &  -- &  -- &  1 &  1 & -- & 1 &  245 ZSM & -- &  -- &  -- &  1 &  -- &  -- &  -- & -- & -- \\
 189 LDE & -- &  -- &  -- &  -- &  -- &  1 &  1 & -- & 1 & \hspace{2mm}  \bf NGTDM \rm  &  &   &   &   &   &   &   &  &  \\
 190 LDHGLE & -- &  -- &  -- &  -- &  -- &  1 &  1 & -- & 1 &  246 Busyness & -- &  -- &  -- &  -- &  -- &  1 &  1 & -- & 1 \\
 191 LDLGLE & -- &  -- &  -- &  -- &  -- &  1 &  1 & -- & 1 &  247 Coarseness & -- &  -- &  -- &  -- &  -- &  1 &  1 & -- & 1 \\
 192 LGLCE & -- &  -- &  -- &  -- &  -- &  1 &  1 & -- & 1 &  248 Complexity & -- &  -- &  -- &  -- &  -- &  1 &  1 & -- & 1 \\
\hspace{2mm}  \bf GLRLM \rm  &  &   &   &   &   &   &   &  &   &  249 Contrast & -- &  -- &  -- &  -- &  -- &  1 &  1 & -- & 1 \\
 193 GLNU & -- &  -- &  -- &  1 &  -- &  1 &  -- & -- &  1 &  250 Strength & -- &  -- &  -- &  -- &  -- &  1 &  1 & -- & 1 \\
 194 GLNUN & -- &  -- &  -- &  1 &  -- &  1 &  -- & -- &  1 & \hspace{2mm}  \bf GLDM \rm  &  &   &   &   &   &   &   &  &  \\
 195 GLV & -- &  -- &  -- &  1 &  -- &  1 &  -- & -- &  1 &  251 SDE & -- &  -- &  -- &  1 &  -- &  1 &  -- & -- & 1 \\
 196 HGLRE & -- &  -- &  -- &  1 &  -- &  1 &  -- & -- &  1 &  252 LDE & -- &  -- &  -- &  1 &  -- &  1 &  -- & -- & 1 \\
   &  &   &   &   &   &   &   &  &   &  253 GLN & -- &  -- &  -- &  1 &  -- &  1 &  -- & -- & 1 \\
   &  &   &   &   &   &   &   &  &   &  254 DN & -- &  -- &  -- &  1 &  -- &  1 &  -- & -- & 1 \\
   &  &   &   &   &   &   &   &  &   &  255 DNN & -- &  -- &  -- &  1 &  -- &  1 &  -- & -- & 1 \\
   &  &   &   &   &   &   &   &  &   &  256 GLV & -- &  -- &  -- &  1 &  -- &  1 &  -- & -- & 1 \\
   &  &   &   &   &   &   &   &  &   &  257 DV & -- &  -- &  -- &  1 &  -- &  1 &  -- & -- & 1 \\
   &  &   &   &   &   &   &   &  &   &  258 DE & -- &  -- &  -- &  1 &  -- &  1 &  -- & -- & 1 \\
   &  &   &   &   &   &   &   &  &   &  259 LGLE & -- &  -- &  -- &  1 &  -- &  1 &  -- & -- & 1 \\
   &  &   &   &   &   &   &   &  &   &  260 HGLE & -- &  -- &  -- &  1 &  -- &  1 &  -- & -- & 1 \\
   &  &   &   &   &   &   &   &  &   &  261 SDLGLE & -- &  -- &  -- &  1 &  -- &  1 &  -- & -- & 1 \\
   &  &   &   &   &   &   &   &  &   &  262 SDHGLE & -- &  -- &  -- &  1 &  -- &  1 &  -- & -- & 1 \\
   &  &   &   &   &   &   &   &  &   &  263 LDLGLE & -- &  -- &  -- &  1 &  -- &  1 &  -- & -- & 1 \\
   &  &   &   &   &   &   &   &  &   &  264 LDHGLE & -- &  -- &  -- &  1 &  -- &  1 &  -- & -- & 1 \\

 \bottomrule
 Total (125) & 13 & 0 & 4 & 67 & 17 & 94 & 78 & 16 & 118 \\
\end{tabular}
\end{tiny}

\vspace{1ex}
\begin{tiny}
{\raggedright
GLCM acronyms:
GLNU = Gray Level NonUniformity,
GLNUN = Gray Level NonUniformity Normalized,
GLV = Gray Level Variance,
HGLRE = High Gray Level Run Emphasis,
LRE = Long Run Emphasis,
LRHGLE = Long Run High Gray Level Emphasis,
LRLGLE = Long Run Low Gray Level Emphasis,
LGLRE = Low Gray Level Run Emphasis,
RE = Run Entropy,
RLNU = Run Length Non Uniformity,
RLNUN = Run Length Non Uniformity Normalized,
RP = Run Percentage,
RV = Run Variance,
Short Run Emphasis,
Short Run High Gray Level Emphasis,
SRLGLE = Short Run Low Gray Level Emphasis.
GLSZM acronyms:
GLNU = Gray Level NonUniformity,
GLNUN = Gray Level NonUniformity Normalized,
GLV = Gray Level Variance,
HGLZE = High Gray Level Zone Emphasis,
LAE = Large Area Emphasis,
LAHGLE = Large Area High Gray Level Emphasis,
LALGLE = Large Area Low Gray Level Emphasis,
LGLZE = Low Gray Level Zone Emphasis,
SZNU = Size Zone NonUniformity,
SZNUN = Size Zone NonUniformity Normalized,
SAE = Small Area Emphasis,
SAHGLE = Small Area High Gray Level Emphasis,
SALGLE = Small AreaLow Gray Level Emphasis,
ZE = Zone Entropy,
ZP = Zone Percentage.

\par}
\end{tiny}
\end{table}


\begin{table}[!hbt]
\centering
\caption {Miscellaneous 2-dimensional features}
\label {table:MISC}
\begin{tiny}
\begin{tabular} {
l  c  c  c  c  c  c  c  c  c  l  c  c  c  c  c  c  c  c  c  l  c  c  c  c  c  c  c  c  c
}
\toprule

Feature & CP &  IM &  MT &  MI & NF &  PR &  RJ & WC & NY \\
\midrule
\hspace{2mm}  \bf Intensity distribution \rm &  &  &  &  &  &  &  &  &  \\
 265 Gabor & -- & -- & -- & -- & -- & -- & -- & 1 & 1 \\
 266 FracAtD & 1 & -- & -- & -- & -- & -- & -- & -- & 1 \\
 267 MeanFrac & 1 & -- & -- & -- & -- & -- & -- & -- & 1 \\
 268 RadialCV & 1 & -- & -- & -- & -- & -- & -- & -- & 1 \\
 269 Zernike & 1 & -- & -- & -- & -- & -- & -- & 1 & 1 \\
\hspace{2mm}  \bf Interdescriptors \rm &  &  &  &  &  &  &  &  &  \\
 270 Num neighbors & 1 & -- & -- & -- & -- & -- & -- & 1 & 1 \\
 271 Percent touching & 1 & -- & -- & -- & -- & -- & -- & -- & 1 \\
 272 1st closest obj number & 1 & -- & -- & -- & -- & -- & -- & -- & -- \\
 273 1st closest dist & 1 & -- & -- & -- & -- & -- & -- & -- & 1 \\
 274 1st closest angle & 1 & -- & -- & -- & -- & -- & -- & -- & 1 \\
 275 2nd closest obj number & 1 & -- & -- & -- & -- & -- & -- & -- & -- \\
 276 2nd closest dist & 1 & -- & -- & -- & -- & -- & -- & -- & 1 \\
 277 2nd closest angle  & 1 & -- & -- & -- & -- & -- & -- & -- & 1 \\
 278 Angle betw neighbors mean & 1 & -- & -- & -- & -- & -- & -- & 1 & 1 \\
 279 Angle betw neighbors med & 1 & -- & -- & -- & -- & -- & -- & -- & 1 \\
 280 Angle betw neighbors STD & 1 & -- & -- & -- & -- & -- & -- & -- & 1 \\
\hspace{2mm}  \bf Skeleton features \rm &  &  &  &  &  &  &  &  &  \\
 281 Number of trunks & 1 & -- & -- & -- & -- & -- & -- & -- & -- \\
 282 Num non-trunk objs & 1 & -- & -- & -- & -- & -- & -- & -- & -- \\
 283 Num branch ends & 1 & -- & -- & -- & -- & -- & -- & -- & -- \\
 284 Total skeleton length & 1 & -- & -- & -- & -- & -- & -- & -- & -- \\
\hspace{2mm}  \bf Image quality \rm &  &  &  &  &  &  &  &  &  \\
 285 IMQ blur FocusScore & 1 & -- & -- & -- & -- & -- & -- & -- & 1 \\
 286 IMQ blur LocalFocusScore & 1 & -- & -- & -- & -- & -- & -- & -- & 1 \\
 287 IMQ blur Correlation & 1 & -- & -- & -- & -- & -- & -- & -- & -- \\
 288 IMQ blur PowLogLogSlope & 1 & -- & -- & -- & -- & -- & -- & -- & 1 \\
 289 IMQ sat PCMax & 1 & -- & -- & -- & -- & -- & -- & -- & 1 \\
 290 IMQ sat PCMin & 1 & -- & -- & -- & -- & -- & -- & -- & 1 \\
 291 IMQ inten Tot I & 1 & -- & -- & -- & -- & -- & -- & -- & -- \\
 292 IMQ inten Mean I & 1 & -- & -- & -- & -- & -- & -- & -- & -- \\
 293 IMQ inten Med I & 1 & -- & -- & -- & -- & -- & -- & -- & -- \\
 294 IMQ inten Std I & 1 & -- & -- & -- & -- & -- & -- & -- & -- \\
 295 IMQ inten MAD I & 1 & -- & -- & -- & -- & -- & -- & -- & -- \\
 296 IMQ inten Max I & 1 & -- & -- & -- & -- & -- & -- & -- & -- \\
 297 IMQ inten Min I & 1 & -- & -- & -- & -- & -- & -- & -- & -- \\
 298 IMQ inten TotArea & 1 & -- & -- & -- & -- & -- & -- & -- & -- \\
 299 IMQ inten TotVol & 1 & -- & -- & -- & -- & -- & -- & -- & -- \\
 300 IMQ inten Scaling & 1 & -- & -- & -- & -- & -- & -- & -- & -- \\
 301 IMQ Sharpness & -- & -- & -- & -- & -- & -- & -- & -- & 1 \\
 302 IMQ Threshold & 1 & -- & -- & -- & -- & -- & -- & -- & -- \\

\bottomrule
Total (38) & 36 & 0 & 0 & 0 & 0 & 0 & 0 & 4 & 20 \\
\end{tabular}
\end{tiny}
\end{table}


\clearpage
\begin{landscape}
\begin{table}[!hbt]
\caption {Volume methods}
\label {table:VOL}
\begin{tiny}
\begin{tabular} {
l		c		c		c		c		c		c		c		c		c		l		c		c		c		c		c		c		c		c		c	
} 
\toprule
Feature	&	CP	& 	IM	& 	MT	& 	MI	&	NF	& 	PR	& 	RJ	&	WC	&	NY	&	Feature	&	CP	& 	IM	& 	MT	& 	MI	&	NF	& 	PR	& 	RJ	&	WC	&	NY	\\
	\midrule

303	Coef of var	&	--	&	--	&	--	&	1	&	--	&	--	&	1	&	--	&	1	&	353	length\_3d\_bb	&	1	&	1	&	--	&	--	&	--	&	--	&	--	&	--	&	1	\\
304	Covered img inten range	&	--	&	--	&	--	&	1	&	--	&	--	&	--	&	--	&	1	&	354	height\_3d\_bb	&	1	&	1	&	--	&	--	&	--	&	--	&	--	&	--	&	1	\\
305	Energy	&	--	&	--	&	--	&	1	&	--	&	1	&	1	&	1	&	1	&	355	feret\_3d\_max	&	--	&	1	&	--	&	--	&	--	&	--	&	--	&	--	&	1	\\
306	Entropy	&	--	&	--	&	1	&	1	&	--	&	1	&	--	&	1	&	1	&	356	feret\_3d\_min	&	--	&	1	&	--	&	--	&	--	&	--	&	--	&	--	&	1	\\
307	Excess kurtosis	&	--	&	--	&	--	&	1	&	--	&	--	&	--	&	--	&	1	&	357	x\_max\_3d	&	--	&	1	&	--	&	--	&	--	&	--	&	--	&	--	&	1	\\
308	Hyperflatness	&	--	&	--	&	--	&	--	&	--	&	--	&	--	&	--	&	1	&	358	y\_max\_3d	&	--	&	1	&	--	&	--	&	--	&	--	&	--	&	--	&	1	\\
309	Hyperskewness	&	--	&	--	&	--	&	--	&	--	&	--	&	--	&	--	&	1	&	359	z\_max\_3d	&	--	&	1	&	--	&	--	&	--	&	--	&	--	&	--	&	1	\\
310	II	&	1	&	--	&	--	&	--	&	--	&	--	&	--	&	1	&	1	&	360	max 3D diam	&	--	&	--	&	--	&	1	&	--	&	1	&	1	&	--	&	1	\\
311	IIE	&	1	&	--	&	--	&	--	&	--	&	--	&	--	&	1	&	1	&	361	texture GLCM	&	1	&	--	&	--	&	1	&	--	&	1	&	1	&	--	&	1	\\
312	IQR	&	--	&	--	&	1	&	1	&	--	&	1	&	1	&	1	&	1	&	362	texture GLDM	&	--	&	--	&	--	&	1	&	--	&	1	&	1	&	--	&	1	\\
313	Kurtosis	&	--	&	--	&	1	&	1	&	--	&	1	&	1	&	1	&	1	&	363	texture GLRLM	&	--	&	--	&	--	&	1	&	--	&	1	&	1	&	--	&	1	\\
314	MAD	&	--	&	--	&	1	&	1	&	--	&	1	&	1	&	1	&	1	&	364	texture GLSZM	&	--	&	--	&	--	&	1	&	--	&	1	&	1	&	--	&	1	\\
315	Max	&	1	&	--	&	1	&	1	&	--	&	1	&	1	&	1	&	1	&	365	texture NGTDM	&	--	&	--	&	--	&	1	&	--	&	1	&	1	&	--	&	1	\\
316	Max IE	&	1	&	--	&	--	&	--	&	--	&	--	&	--	&	1	&	1	&	366	texture NGLDM	&	--	&	--	&	--	&	1	&	--	&	--	&	--	&	--	&	--	\\
317	MD	&	1	&	--	&	--	&	--	&	--	&	--	&	--	&	1	&	1	&	367	texture GLDZM	&	--	&	--	&	--	&	1	&	--	&	--	&	--	&	--	&	--	\\
318	Mean	&	1	&	--	&	1	&	1	&	--	&	1	&	1	&	1	&	1	&	368	Raw moments	&	--	&	--	&	--	&	--	&	--	&	--	&	--	&	--	&	1	\\
319	Mean IE	&	1	&	--	&	--	&	--	&	--	&	--	&	--	&	--	&	1	&	369	Weighted raw moments	&	--	&	--	&	--	&	--	&	--	&	--	&	--	&	--	&	--	\\
320	Median	&	--	&	--	&	1	&	1	&	--	&	1	&	1	&	1	&	1	&	370	Central moments	&	--	&	--	&	--	&	--	&	--	&	--	&	--	&	--	&	1	\\
321	Median abs dev	&	--	&	--	&	--	&	1	&	--	&	--	&	1	&	--	&	1	&	371	Weighted central moments	&	--	&	--	&	--	&	--	&	--	&	--	&	--	&	--	&	--	\\
322	Min	&	1	&	--	&	1	&	1	&	--	&	1	&	1	&	1	&	1	&	372	Normalized central moments	&	--	&	--	&	--	&	--	&	--	&	--	&	--	&	--	&	--	\\
323	Min IE	&	1	&	--	&	--	&	--	&	--	&	--	&	--	&	1	&	1	&	373	Normalized raw moments	&	--	&	--	&	--	&	--	&	--	&	--	&	--	&	--	&	--	\\
324	Mode	&	--	&	--	&	1	&	1	&	--	&	--	&	--	&	1	&	1	&	374	Hu invariants	&	--	&	--	&	--	&	--	&	--	&	--	&	--	&	--	&	1	\\
325	Mode probability	&	--	&	--	&	--	&	1	&	--	&	--	&	--	&	--	&	1	&	375	Weighted Hu invariants	&	--	&	--	&	--	&	--	&	--	&	--	&	--	&	--	&	--	\\
326	Percentiles	&	--	&	--	&	1	&	1	&	--	&	1	&	1	&	1	&	1	&	376	neigh/NumberOfNeighbors	&	1	&	--	&	--	&	--	&	--	&	--	&	--	&	--	&	1	\\
327	Quantile coef of disp	&	--	&	--	&	--	&	1	&	--	&	--	&	1	&	--	&	1	&	377	neigh/PercentTouching	&	1	&	--	&	--	&	--	&	--	&	--	&	--	&	--	&	1	\\
328	Range	&	--	&	--	&	1	&	1	&	--	&	1	&	1	&	1	&	1	&	378	neigh/FirstClosestObjectNumber	&	1	&	--	&	--	&	--	&	--	&	--	&	--	&	--	&	--	\\
329	RMAD	&	--	&	--	& 	--	& 	1	&	--	&	1	& 	--	&	1	&	1	&	379	neigh/FirstClosestDistance	&	1	&	--	&	--	&	--	&	--	&	--	&	--	&	--	&	1	\\
330	Rmean	&	--	&	--	&	--	&	1	&	--	&	--	&	--	&	--	&	1	&	380	neigh/SecondClosestObjectNumber	&	1	&	--	&	--	&	--	&	--	&	--	&	--	&	--	&	--	\\
331	RMS	&	--	&	--	&	1	&	1	&	--	&	1	&	1	&	1	&	1	&	381	neigh/SecondClosestDistance	&	1	&	--	&	--	&	--	&	--	&	--	&	--	&	--	&	1	\\
332	SE	&	--	&	--	&	1	&	--	&	--	&	--	&	1	&	1	&	1	&	382	neigh/AngleBetweenNeighbors	&	1	&	--	&	--	&	--	&	--	&	--	&	--	&	--	&	1	\\
333	Skewness	&	--	&	--	&	1	&	1	&	--	&	1	&	1	&	1	&	1	&	383	imq/blur/FocusScore	&	1	&	--	&	--	&	--	&	--	&	--	&	--	&	--	&	--	\\
334	std	&	1	&	--	&	1	&	1	&	--	&	1	&	1	&	1	&	1	&	384	imq/blur/LocalFocusScore	&	1	&	--	&	--	&	--	&	--	&	--	&	--	&	--	&	--	\\
335	std IE	&	1	&	--	&	--	&	--	&	--	&	--	&	--	&	--	&	1	&	385	imq/blur/Correlation	&	1	&	--	&	--	&	--	&	--	&	--	&	--	&	--	&	--	\\
336	std biased	&	--	&	--	&	--	&	1	&	--	&	--	&	--	&	--	&	1	&	386	imq/blur/PowerLogLogSlope	&	1	&	--	&	--	&	--	&	--	&	--	&	--	&	--	&	--	\\
337	Uniformity	&	--	&	--	&	--	&	1	&	--	&	1	&	--	&	--	&	1	&	387	imq/sat/PercentMax	&	1	&	--	&	--	&	--	&	--	&	--	&	--	&	--	&	--	\\
338	Uniformity PIU	&	--	&	--	&	--	&	--	&	--	&	--	&	--	&	--	&	1	&	388	imq/sat/PercentMin	&	1	&	--	&	--	&	--	&	--	&	--	&	--	&	--	&	--	\\
339	Variance	&	--	&	--	&	1	&	1	&	--	&	1	&	1	&	1	&	1	&	389	imq/inten/TotalIntensity	&	1	&	--	&	--	&	--	&	--	&	--	&	--	&	--	&	--	\\
340	Variance (biased)	&	--	&	--	&	--	&	1	&	--	&	--	&	--	&	--	&	1	&	390	imq/inten/MeanIntensity	&	1	&	--	&	--	&	--	&	--	&	--	&	--	&	--	&	--	\\
341	WCX	&	--	&	--	&	1	&	--	&	--	&	--	&	--	&	1	&	1	&	391	imq/inten/MedIntensity	&	1	&	--	&	--	&	--	&	--	&	--	&	--	&	--	&	--	\\
342	WCY	&	--	&	--	&	1	&	--	&	--	&	--	&	--	&	1	&	1	&	392	imq/inten/StdIntensity	&	1	&	--	&	--	&	--	&	--	&	--	&	--	&	--	&	--	\\
343	volume	&	1	&	1	&	--	&	1	&	--	&	1	&	1	&	--	&	1	&	393	imq/inten/MADIntensity	&	1	&	--	&	--	&	--	&	--	&	--	&	--	&	--	&	--	\\
344	volume\_bb	&	1	&	--	&	--	&	--	&	--	&	--	&	--	&	--	&	1	&	394	imq/inten/MaxIntensity	&	1	&	--	&	--	&	--	&	--	&	--	&	--	&	--	&	--	\\
345	volume\_convexhull	&	--	&	1	&	--	&	--	&	--	&	--	&	--	&	--	&	1	&	395	imq/inten/MinIntensity	&	1	&	--	&	--	&	--	&	--	&	--	&	--	&	--	&	--	\\
346	convex area	&	1	&	--	&	--	&	--	&	--	&	--	&	--	&	--	&	1	&	396	imq/inten/TotalArea	&	1	&	--	&	--	&	--	&	--	&	--	&	--	&	--	&	--	\\
347	surface\_area	&	1	&	1	&	--	&	--	&	--	&	--	&	--	&	--	&	1	&	397	imq/inten/TotalVol	&	1	&	--	&	--	&	--	&	--	&	--	&	--	&	--	&	--	\\
348	D\_volume\_equivalent	&	--	&	1	&	--	&	--	&	--	&	--	&	--	&	--	&	1	&	398	imq/inten/Scaling	&	1	&	--	&	--	&	--	&	--	&	--	&	--	&	--	&	--	\\
349	D\_surfacearea\_equivalent	&	--	&	1	&	--	&	--	&	--	&	--	&	--	&	--	&	1	&	399	imq/Threshold	&	1	&	--	&	--	&	--	&	--	&	--	&	--	&	--	&	--	\\
350	extent	&	1	&	--	&	--	&	--	&	--	&	--	&	--	&	--	&	--	&			&		&		&		&		&		&		&		&		&		\\
351	centroid	&	1	&	1	&	--	&	--	&	--	&	--	&	--	&	--	&	1	&			&		&		&		&		&		&		&		&		&		\\
352	width\_3d\_bb	&	1	&	1	&	--	&	--	&	--	&	--	&	--	&	--	&	1	&			&		&		&		&		&		&		&		&		&		\\
		
\bottomrule
Total (97) & 45 & 14 & 18 & 36 & 0 & 24 & 25 & 25 & 70 \\
\end{tabular}																																			
\end{tiny}																																				
\vspace{1ex}																																			
\begin{tiny}																																				
{\raggedright																																			
\par}																																				
\end{tiny}																																				

\end{table}
\end{landscape}
\clearpage


\begin{table}[!hbt]
\centering
\caption {Benchmarks on Tissuenet and Decathlon datasets by feature group. All values are in seconds. NY$_{\textrm{U}}$ refers to the Nyxus untargeted profile (all features enabled), and NY$_{\textrm{T}}$ refers to the Nyxus targeted profile (optimized feature subset). These raw timing values correspond to the fold-change comparisons shown in Figure \ref{fig:PERF}.}
\label {table:BENCH}
\begin{tabular} { l		c		c		c		c		c		c		c 		c	 		c c 		} 
\toprule

			&	CP$^1$	& 	IM$^2$	& 	MT$^3$	& 	MI$^4$	&	NF$^5$	& 	PR$^6$	& 	RJ$^7$	&	WC$^8$	& NY$_{\textrm{U}}^9$ & NY$_{\textrm{T}}^{10}$	\\
\midrule
\hspace{2mm}\textbf{tissuenet}	&		&		&		&		&		&		&		&		&			&			\\
intensity	& 211	& --	& 110	&446726 & 116	& 152	&418842	& 82	& 65		& 4			\\
shape		& 1516	& 5892	& 612	& --	& 54	& --	& 15613	& 106	& 86		& 35		\\
texture 	& 5334	& --	& 740	&450056	& 96	& 875	& 62034	& 58	& 153		& 41		\\
misc. 		& 1267	& 15409	& --	& --	& --	& --	& --	& 63	& 160		& 129		\\
\hspace{2mm}\textbf{decathlon}	&		&		&		&		&		&		&		&		&			&			\\	
intensity	& 25261	& --	& 5720	& 2577	& 5657	&113150	& 14790	& 4541	& 1317		& 1327		\\
shape		&181418	& 43047	& 6764	& --	& 6076	&109522	& 19143	& 8343	& 1975		& 1982		\\
texture 	&638068	& --	& 6436	& 2618	& 5680	&112865	& 17123	& 22139	& 3218		& 1789		\\
misc. 		&151519	& --	& --	& --	& --	& --	& --	& --	& 2156		& 2167		\\
\bottomrule
\end{tabular}

\vspace{1ex}
\begin{tiny}
{\raggedright 
1 -- CellProfiler, 
2 -- Imea, 
3 -- Matlab R2022b Image processing toolkit,
4 -- MITK,
5 -- NIST Feature2DJava, 
6 -- PyRadiomics,
7 -- RadiomicsJ,
8 -- WindCharm,
9 -- Nyxus (untargeted feature extraction),
10 -- Nyxus (targeted feature extraction)
\par}
\end{tiny}
\end{table}

\clearpage
\begin{landscape}
\begin{tiny}
\begin{longtable}{p{2cm} | p{5cm} | p{3cm} | p{3cm} | p{3cm}}
\caption {Criteria behind the functionality heatmap (from Figure \ref{fig:FUNCTIONALITY_HEATMAP})}
\label {table:FUNC_HEATMAP_CRITERIA} \\
\toprule
	Category & Remark  & Green	& 	Blue	& 	Red	\\
	\midrule
\endfirsthead
\multicolumn{5}{c}{\tablename\ \thetable{} -- continued from previous page} \\
\toprule
	Category & Remark  & Green	& 	Blue	& 	Red	\\
	\midrule
\endhead
\midrule
\multicolumn{5}{r}{Continued on next page} \\
\endfoot
\bottomrule
\endlastfoot
Out of Core Support 
	& Can the Method calculate all (or the majority) of its features if an image or ROI is larger than available RAM? 
	& $>90\%$
	& $10\% \leqslant x \leqslant 90\%$
	& $< 10\%$
	\\
	\hline
\# Of Feature Classes 
	& Count of unique measurements extracted by each Method. Supplemental Tables \ref{table:INTEN}-\ref{table:VOL} display all feature classes, each class is given a unique number in these tables. Of note is that often within a given feature depending on feature hyperparameters, many different measurements can be made. For example, Zernike polynomials measure the intensity distribution from each object’s center to its boundary within a set of bins, across its radial and azimuthal degrees and therefore can have unlimited Zernike polynomial measures for a single feature class. However, for this manuscript Zernike would represent only one feature class 
	& $>90\%$
	& $10\% \leqslant x \leqslant 90\%$
	& $< 10\%$ \\
	\hline
Next Gen. Output Format 
	& Method supports both in memory representation and saving to disk of measured feature classes in indexed columnar formats such as Apache Arrow, Feather, and/or Parquet 
	& Supports $>1$ such format
	& Supports 1 such format
	& Does not natively support output of feature values in an indexed columnar format
	\\ 
	\hline
GPU Support 
	& Method supports calculation of features using a GPU 
	& Supports calculating > 1 feature classes on GPU
	& Supports calculating 1 feature class on GPU
	& Does not support calculation of any feature classes on GPU
	\\
	\hline
Soft (Tunable) Features 
	& Method supports users setting hyperparameters for features that have hyperparameters 
	& $>90\%$
	& $10\% \leqslant x \leqslant 90\%$
	& $< 10\%$
	\\
	\hline
IBSI Compliant 
	& Method can be automatically configured to output feature measurement values that fall within accepted ranges on IBSI benchmark datasets for feature classes within the IBSI standard 
	& $>90\%$
	& $10\% \leqslant x \leqslant 90\%$
	& $< 10\%$
	\\
	\hline
OME Tiff and NGFF Support 
	& Method can load and process images in the OME.tiff and OME.ngff (zarr) format
	& Method can open both file formats and process them without the user having to install or utilize other Methods/tools
	& Method can open both file formats and process them using third party Methods/tools, without converting the files into another format
	& Method cannot open both file formats without converting them to another file format
	\\
	\hline
DICOM Support  
	& Method can load and process images in the DICOM format
	& Method can open files in this format and process them without the user having to install or utilize other Methods/tools
	& Method can open files in this format and process them using third party Methods/tools, without converting the files into another format
	& Method cannot open files in this format without converting them to another file format
	\\
	\hline
ROI/FOV  
	& ROI means that a Method can extract feature class measurements from an image and a corresponding mask together producing measurements per unique region in the mask. FOV means that a Method can extract feature class measurements from an image file without the need of a mask, producing one 	set of measurements per image file. Of note is that a Method is not given credit for FOV support if the user must submit a mask with an image that covers the entire area of the image
	& Both –- Supports both ROI and FOV measurements
	& 
	& ROI or FOV -- Only supports either ROI measurements or FOV measurements, not both
	\\
	\hline
3D Feature Classes  
	& Count of feature classes a Method supports that can be measured on volumetric (3D) regions. Of note is that a Method that supports extracting features on individual 2D slices of 3D volumes is not given credit for supporting 3D feature classes 
	& $>90\%$
	& $10\% \leqslant x \leqslant 90\%$
	& $< 10\%$
	\\
	\hline
GUI  
	& The maintainer of the Method has developed a GUI that enables the use of the Method without programming or command line use
	& Method can be accessed from a GUI that the maintainer of the tool has developed OR the GUI for the Method can be accessed from a platform that is developed and maintained by the same organization that develops and maintains the Method
	& (via X) Method can be accessed from a GUI that the maintainer of the tool has developed in a platform that is not developed and maintained by the same organization that develops and maintains the Method
	& Method cannot be accessed from a GUI that the maintainer of the tool has developed
	\\
	\hline
Standard Workflow Tool Definition  
	& Method is made available as an open standard workflow tool/step by the maintainer of the Method. Examples include: Common Workflow Language Command Line Tool (CWL-CLT), Workflow Definition Language (WDL), Nextflow (NF), Interoperable Computational Tool (ICT), Web Image Processing Pipeline Plugin (WIPP Plugin), etc
	& X -- Method is made available by the maintainer of the Method as more than one open standard workflow language tool/step
	& X (Partial) –– Method is made available by the maintainer of the Method as a single open standard workflow language tool/step
	& No -- Method is not made available by the maintainer of the Method as an open standard workflow language tool/step
	\\
	\hline
OCI Container  
	& Method is made available as an Open Container Initiative (OCI) compliant container by the maintainer of the tool. Examples include: Docker container, Aptainer, Podman container, Singularity container, etc
	& Yes –- Method is made available by the maintainer of the Method in at least one OCI compliant container format
	&
	& No –- Method is not made available by the maintainer of the Method in at least one OCI compliant container format
	\\
	\hline
Command Line Tool  
	& Method is made available as a command line tool by the maintainer of the tool
	& Yes –- Method is made available by the maintainer of the Method as a command line tool. Users do not need to compile their own command line tool
	&
	& No –- Method is not made available by the maintainer of the Method as a command line tool
	\\
	\hline
Image Formats Supported  
	& Count of the number of input image formats (or the format if only one) the Method supports along with a citation
	& Supports $> 90\%$ of all formats supported across all libraries
	& Supports loading $>1$ image format
	& Supports loading a single format
	\\
	\hline
Licensing  
	& Description of the license type associated with the Method
	& License name - Green – All BSD, MIT, and Apache v2.0 as well as unlicensed Methods, e.g. most permissive
	& License name -- All copyleft licenses (for example GPL and AGPL), e.g. license restrictions
	& License name -- Proprietary license for the software, e.g. most restrictive
	\\
	\hline
Open Source  
	& Method maintainers have made the source code (not just binaries) publicly available for inspection and validation
	& Yes -- Method maintainers provide access to all source code and its documentation in a public repository(s)
	& Semi -- Method maintainers provide access to some pieces of the code and its documentation in a public repository(s)
	& No -- Method maintainers provide no access to source code and/or its documentation in a public repository(s)
	\\
	\hline
Languages  
	& A list of languages the Method maintainers have made the Method available/accessible in
	& & & \\
\end{longtable}
\end{tiny}
\end{landscape}
\clearpage

\end{document}